\documentclass[10pt,twocolumn,letterpaper]{article}
\PassOptionsToPackage{table,dvipsnames}{xcolor}
\usepackage{iccv}
\usepackage{cuted}
\usepackage{caption}
\usepackage{amsmath,amssymb,bm}
\usepackage{booktabs,multirow,makecell}
\usepackage{xcolor,colortbl}
\usepackage{graphicx}
\usepackage{enumitem}
\usepackage{xspace}
\newcommand{\ArtMesh}{ArtMesh}
\definecolor{ourscolor}{RGB}{220,230,240}
\newcommand{\our}[1]{\cellcolor{ourscolor}#1}

\usepackage{tikz}
\usepackage[skins]{tcolorbox}
\definecolor{cMesh}{HTML}{1F7A74}
\definecolor{cCons}{HTML}{6F5BA8}
\definecolor{cBench}{HTML}{C9841C}
\definecolor{cIdea}{HTML}{4A6B8A}
\newtcolorbox{ideabox}[1]{enhanced,colback=#1!14!white,colframe=#1!80!black,boxrule=0.5pt,arc=2.5pt,left=4pt,right=4pt,top=2.5pt,bottom=2.5pt,boxsep=0pt,before skip=4pt plus 1pt,after skip=4pt plus 1pt,fontupper=\small}
\newcommand{\gapbox}[3]{\begin{ideabox}{#1}\textbf{#2} #3\end{ideabox}}
\newcommand{\mk}[2]{\tikz[baseline=(m.base)]{\node[circle,draw=#2!80!black,fill=#2!20!white,line width=0.45pt,inner sep=0pt,minimum size=1.3em,font=\sffamily\bfseries\scriptsize,text=#2!50!black](m){#1};}}
\newcommand{\mM}{\mk{M}{cMesh}\,}
\newcommand{\mC}{\mk{C}{cCons}\,}
\newcommand{\mB}{\mk{B}{cBench}\,}

\definecolor{linkblue}{rgb}{0.21,0.49,0.74}
\usepackage[pagebackref,breaklinks,colorlinks,allcolors=linkblue]{hyperref}

\title{ArtMesh: Part-Aware Articulated Mesh Fields with Motion-Consistent Dynamics}

\author{
\textbf{Sylvia Yuan}\textsuperscript{1,*} \quad
\textbf{Dan Wang}\textsuperscript{2,*} \quad
\textbf{Ravi Ramamoorthi}\textsuperscript{1} \quad
\textbf{Xinrui Cui}\textsuperscript{3,\dag}
\\[4pt]
\textsuperscript{1}University of California San Diego \quad
\textsuperscript{2}University of Arizona \quad
\textsuperscript{3}University of North Texas
\\[2pt]
{\small \textsuperscript{*}Equal contribution \quad \textsuperscript{\dag}Corresponding author}
}

\begin{document}
\maketitle
\begin{strip}
\centering
\includegraphics[width=\linewidth]{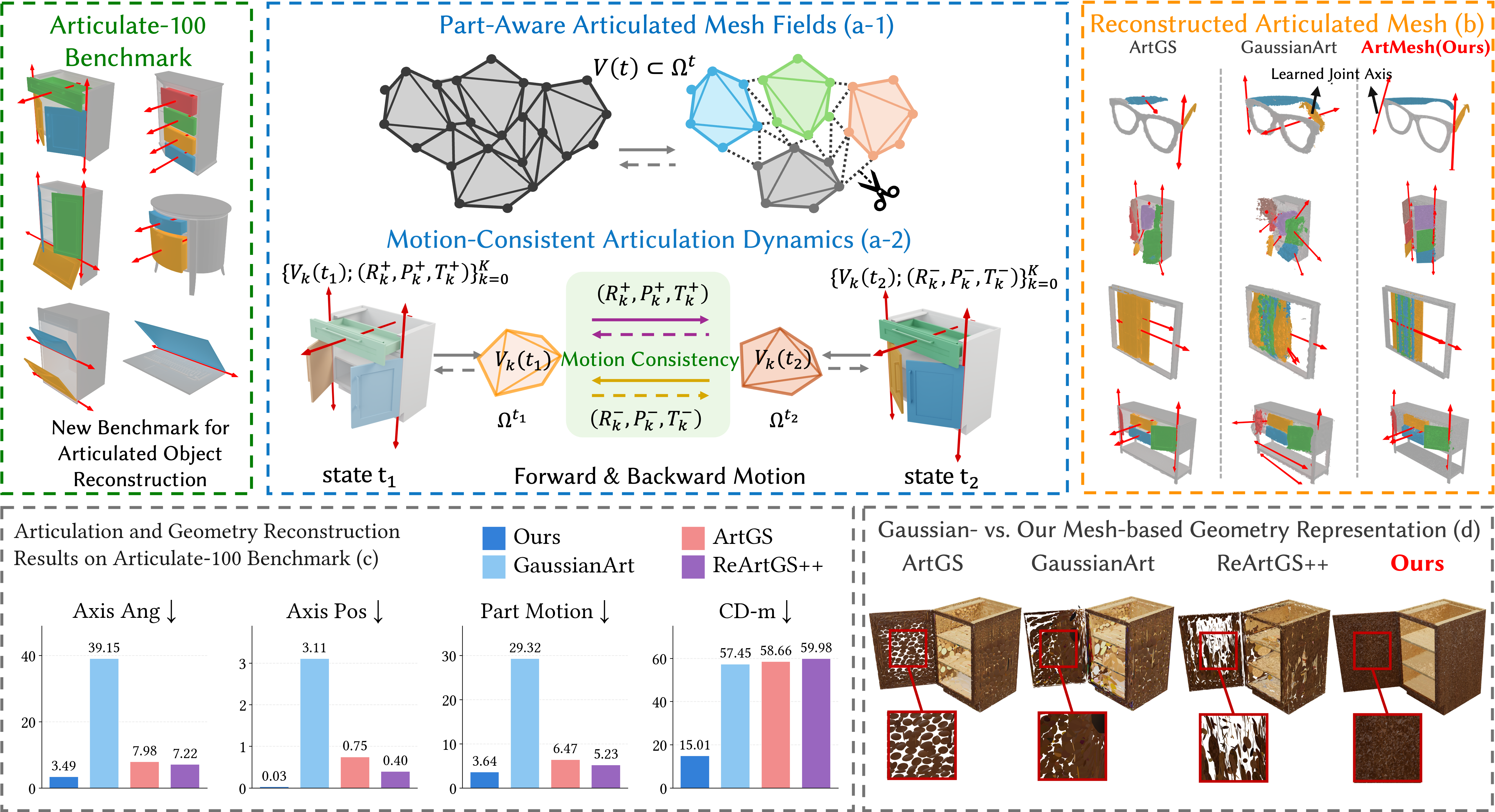}
\captionsetup{hypcap=false}
\captionof{figure}{\textbf{\ArtMesh{} reconstructs articulated objects as part-aware connected triangle meshes with per-part rigid motion.} \textbf{(a)} From multi-view observations at two states, we jointly recover a part-aware mesh field via per-part restricted Delaunay remeshing, \protect\mM, and a motion-consistent articulation field, \protect\mC. \textbf{(b)} Start-state meshes transformed to the end state by the learned articulation, with predicted joint axes (red arrows; lengths not meaningful); \ArtMesh{} produces these meshes directly, without post-hoc TSDF fusion. \textbf{(c)} Quantitative comparison on \emph{Articulate-100}, \protect\mB; in (b) and (c) all methods use ground-truth part masks. \textbf{(d)} Baselines output unstructured Gaussians; \ArtMesh{} outputs a connected, opaque triangle mesh.}
\captionsetup{hypcap=true}
\label{fig:teaser}
\end{strip}

\begin{abstract}
We present \ArtMesh, a mesh-native method for reconstructing articulated objects explicitly as connected triangle meshes with per-part rigid motion from multi-view observations in start and end states. Existing 3D Gaussian Splatting pipelines for articulated reconstruction inherit the unstructured point-based geometry of their splatting base, which provides no surface topology for reasoning about part boundaries or enforcing motion consistency along the object's connectivity. \ArtMesh{} instead builds on a mesh-based differentiable rendering backbone, enabling part-aware dynamics to act directly on the structured topology.
To make the topology compatible with articulation, we introduce part-aware restricted Delaunay remeshing, producing connected submeshes whose triangles do not cross semantic part boundaries.
The dynamic mesh field then optimizes articulation using bidirectional Vertex-wise Motion Consistency on transported mesh vertices and Pixel-wise Motion Consistency on rendered images.
We introduce Articulate-100, a new benchmark of 100 articulated objects spanning 16 PartNet-Mobility categories. On this benchmark, given ground-truth part masks, \ArtMesh{} achieves the best aggregate joint-axis estimation and part-level geometric reconstruction among prior 3DGS-based pipelines, with the largest gains on objects with many movable parts.

\end{abstract}

\section{Introduction}
\label{sec:intro}

Reconstructing articulated objects from images is a core problem in building digital twins for robotics, embodied AI, AR/VR, and physical simulation. A useful reconstruction must deliver three things: part-level geometry, the joint motion of each movable part, and an explicit surface asset that downstream systems can consume. Existing pipelines fall short of this goal in two ways. Implicit neural representations, e.g.\ PARIS~\cite{paris}, render well but optimize a volumetric field, so a mesh must be extracted post hoc, losing sharp boundaries and joint topology. Gaussian-splatting pipelines such as ArtGS~\cite{artgs} and GaussianArt~\cite{gaussianart} inherit unstructured point-based geometry without triangle connectivity or explicit part boundaries (Fig.~\ref{fig:teaser}(d); full surface comparison in Fig.~\ref{fig:more-mesh-quality} in the appendix). A mesh must then be recovered by post-hoc TSDF fusion, which introduces holes, smoothing, and ambiguous connectivity near joints, where a single wrong connection across a drawer seam corrupts both the recovered geometry and the estimated motion. We discuss both families in detail in Sec.~\ref{sec:related}.
\gapbox{cMesh}{Challenge 1 (no explicit part-aware surface).}{Implicit fields and Gaussian clouds carry no triangle connectivity and no part boundaries, so a mesh has to be recovered after training, and one wrong connection across a drawer seam corrupts both the geometry and the motion.}

A second challenge is motion learning: image-space losses supervise 3D kinematic parameters (joint axes, pivots, translations) only indirectly, and small photometric errors can correspond to large joint-axis errors. Prior methods relate articulation states via 2D feature matching~\cite{digitaltwinart}, canonical Gaussian matching~\cite{artgs}, or soft mixtures of part-level motions~\cite{gaussianart}, cues that are fragile on weakly textured parts; without an explicit part-aware surface, supervision leaks across neighboring components and different rigid parts share incorrect motion.
\gapbox{cCons}{Challenge 2 (images constrain joints weakly).}{A small photometric error can hide a large joint-axis error, and the 2D feature matches used to bridge the gap are unreliable on the weakly textured, repetitive parts of furniture.}

A third challenge is evaluation. Most articulated-object benchmarks are two-part (e.g., PARIS) or target perception and manipulation (e.g., GAPartNet~\cite{gapartnet}); MPArt-90~\cite{gaussianart} is a recent multi-part exception.
\gapbox{cBench}{Challenge 3 (benchmarks stop at two parts).}{Few benchmarks go beyond one movable part, so the regime where topology, part separation and motion consistency matter most is not measured.}

We present \textbf{ArtMesh}, a mesh-native method for reconstructing articulated objects from multi-view RGB-D observations captured at two articulation states (Fig.~\ref{fig:teaser}). ArtMesh represents an object as explicit part-aware triangle meshes linked by per-part rigid motions (Fig.~\ref{fig:teaser}(a)). Unlike Gaussian-based pipelines that output ellipsoid primitive-based surfaces, our mesh surface optimized during training is the same surface exported at test time (Fig.~\ref{fig:teaser}(b,d)).

The key idea of ArtMesh is to make geometry, topology, and articulation mutually constrained (Fig.~\ref{fig:arch}). Building on mesh-native differentiable rasterization~\cite{meshsplatting}, we attach geometry, appearance, opacity, and part information to mesh vertices, harden the part assignments, and \protect\mM run part-aware restricted Delaunay remeshing (Fig.~\ref{fig:arch}(a)) within each semantic part, producing connected per-part submeshes with no cross-part triangles, so each movable component transforms as a rigid mesh. An articulation dynamic field (Fig.~\ref{fig:arch}(b)) assigns each part a rigid motion between the two states, with the backward motion the analytic inverse of the forward one. \protect\mC The articulation is optimized with motion consistency learning (Sec.~\ref{method:consistency}): vertex-wise consistency transports each vertex through its part motion and compares it with the surface of the same part in the other state, while pixel-wise consistency renders the transported mesh from target-state cameras and compares with the target images.
\gapbox{cIdea}{Key idea.}{Geometry, topology and articulation constrain one another. Per-part connectivity turns every movable component into a rigid mesh, and that rigid structure is what the bidirectional motion-consistency losses supervise, so the joints are estimated on the same surface that is exported.}

\mB To evaluate articulated reconstruction beyond simple two-part examples, we introduce Articulate-100 (Fig.~\ref{fig:teaser} and Sec.~\ref{sec:articulate100}), a benchmark of 100 articulated objects sampled from PartNet-Mobility~\cite{sapien} across 16 categories and a wide range of part counts. This benchmark stresses multi-part objects where topology, part separation, and motion consistency become increasingly important. On Articulate-100 with ground-truth part masks, ArtMesh achieves the best aggregate performance on joint-axis estimation and part-level geometry reconstruction, and shows the strongest gains on objects with four or more parts (Fig.~\ref{fig:teaser}(c)).

Our contributions are:
\begin{itemize}[topsep=0pt, itemsep=0pt, partopsep=0pt, parsep=0pt, leftmargin=*]
\item \mM\textbf{Part-aware articulated mesh representation (Sec.~\ref{sec:part_aware_mesh_fields})}: explicit per-part triangle submeshes linked by rigid joints, yielding a directly usable mesh asset without post-hoc conversion; part-aware restricted Delaunay remeshing removes cross-part triangles that would deform incorrectly under rigid motion.
\item \mC\textbf{Motion-consistency articulation learning (Sec.~\ref{method:consistency})}: we optimize per-part rigid motion by proposed bidirectional vertex-wise and pixel-wise motion consistency, with the backward motion implemented as the analytic inverse of the forward motion.
\item \mB\textbf{Articulate-100 benchmark (Sec.~\ref{sec:articulate100})}: a 100-object benchmark spanning 16 PartNet-Mobility~\cite{sapien} categories and diverse part counts, on which ArtMesh, given ground-truth part masks, achieves the best aggregate joint estimation and part-level geometry (Fig.~\ref{fig:teaser}(c)), with larger gains on objects with more parts.
\end{itemize}

\section{Related Work}
\label{sec:related}
\textbf{Geometric representations for articulated objects.}
Implicit methods built on NeRF~\cite{nerf} or SDFs~\cite{neus}, A-SDF~\cite{asdf}, CLA-NeRF~\cite{cla-nerf}, NARF~\cite{narf}, PARIS~\cite{paris}, DigitalTwinArt\-~\cite{digitaltwinart}, Ditto~\cite{ditto}, LEIA~\cite{leia}, Articulate-Your-NeRF~\cite{articulate-nerf}, output volumetric fields needing lossy post-hoc isosurfacing decoupled from optimization. Following 3D Gaussian Splatting~\cite{3dgs}, a line of work~\cite{artgs,gaussianart,splart,part2gs,articulated-gs,reartgs,reartgspp,screwsplat,structuregs} replaces this field with explicit Gaussians; some add surface or structural constraints (e.g., ReArtGS~\cite{reartgs} regularizes the Gaussian opacity field with an unbiased SDF, its follow-up ReArtGS++~\cite{reartgspp} instead uses planar Gaussians with a temporal geometry constraint across articulation states, and StructureGS~\cite{structuregs}, the closest in spirit, adds per-part oriented boxes and contact), but all inherit unstructured point-based geometry, gaps and ill-defined surfaces at part boundaries, no connectivity topology, so mesh recovery again needs lossy post-processing. \ArtMesh{} departs from both by reconstructing a connected triangle mesh end-to-end.

\textbf{Mesh-native differentiable reconstruction.}
A separate line renders explicit surfaces: mesh rasterizers~\cite{n3dmr,softras} established mesh-level gradients, and later work converts splatting to meshes~\cite{2dgs,gof,rade-gs,sugar,milo} or renders mesh-like primitives, Triangle Splatting~\cite{triangle-splatting} (unconnected soup) and MeshSplatting~\cite{meshsplatting}, which yields connected, near-manifold meshes via a restricted Delaunay step. \ArtMesh{} adopts its backbone and runs the Delaunay step \emph{per part} to respect piecewise-rigid structure.

\textbf{Feedforward and language-driven articulation estimation.}
Orthogonal approaches treat articulation as a prediction problem: regressing motion parameters, masks, actionability, or URDF assets from few images or point clouds~\cite{shape2motion,captra,where2act,gapartnet,sage,real2code,urdformer,larm,opd,opdmulti,singapo,dreamart,partrm}, or leveraging vision-language models to generate URDFs and affordances~\cite{articulate-anything,a3vlm,manipllm,urdfanything}. These are fast at inference but largely bounded by their training data, with geometry often coarse or retrieved from a mesh library. \ArtMesh{} targets the per-object regime, where fidelity matters more than generalization and the surface is recovered from observations.

\section{Method}
\label{sec:method}

\begin{figure*}[t]
    \centering
    \includegraphics[width=\linewidth]{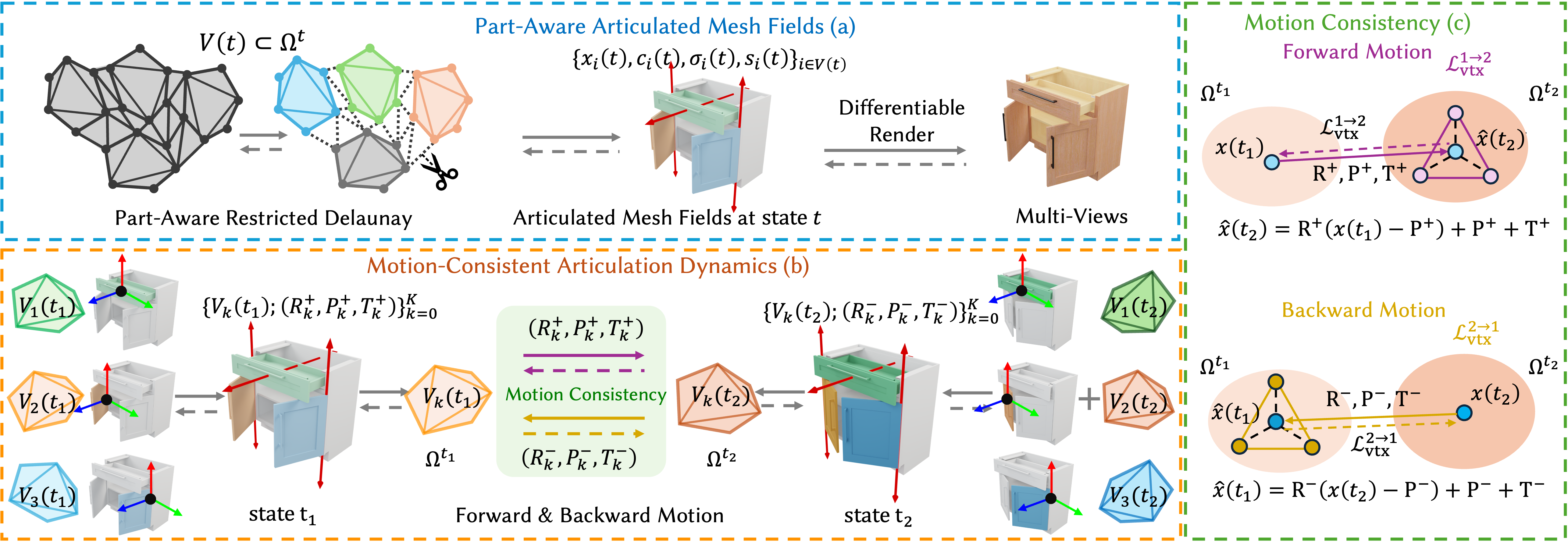}
    \caption{\textbf{Overview of \ArtMesh.} From multi-view RGB-D observations at two states $t_1,t_2\in\{0,1\}$, we reconstruct a pair of part-aware triangle meshes in correspondence and the per-part rigid articulation relating them. \protect\mM The \emph{Part-Aware Mesh Field} (Sec.~\ref{sec:part_aware_mesh_fields}) partitions each state's mesh into per-part clusters via part-aware restricted Delaunay triangulation (Fig.~\ref{fig:render}(a)). The \emph{Articulation Dynamic Field} assigns each part a forward rigid motion $(R_k^{+},P_k^{+},T_k^{+})$ and its analytic inverse. \protect\mC The \emph{Articulation-aware Motion Consistency Loss} (Sec.~\ref{method:consistency}) transports each vertex through the articulation and compares its position, color and opacity against the closest same-part triangle in the other state, in both directions.}
\label{fig:arch}
\end{figure*}

\textbf{Framework overview.}
As illustrated in Fig.~\ref{fig:arch}, ArtMesh reconstructs an articulated object from multi-view observations captured at two states $t_1,t_2\in\{0,1\}$. Given multi-view RGB-D images at two articulation states, we reconstruct a connected part-aware triangle mesh together with one rigid transform per movable part (joint axis, pivot, and translation) linking the two states, exportable as a URDF asset (Sec.~\ref{experiments:simulation}).

\textbf{ArtMesh main components.} \textbf{1)} \protect\mM The \textit{Part-Aware Articulated Mesh Field} (Fig.~\ref{fig:arch}(a)) represents each state-$t$ surface as a triangle mesh with per-vertex position, radiance, opacity, and part weights, hardened and remeshed independently within each semantic part so no triangle crosses a part boundary; \textbf{2)} The \textit{Articulation Dynamic Field} (Fig.~\ref{fig:arch}(b)) assigns each part a forward rigid transform $(R_k^+, P_k^+, T_k^+)$ from $t_1$ to $t_2$ and its analytic inverse, linking the part meshes across states with a single consistent motion; \textbf{3)} \protect\mC \textit{Articulation-Aware Motion Consistency Learning} (Fig.~\ref{fig:arch}(c)) optimizes these motions bidirectionally: a vertex-wise loss transports vertices through the part motion and matches them to the same-part target surface, and a pixel-wise loss renders the transported mesh from target-state cameras against the observed images.

\subsection{Part-Aware Articulated Mesh Field}
\label{sec:part_aware_mesh_fields}
\begin{figure*}[t]
    \centering
    \includegraphics[width=\linewidth]{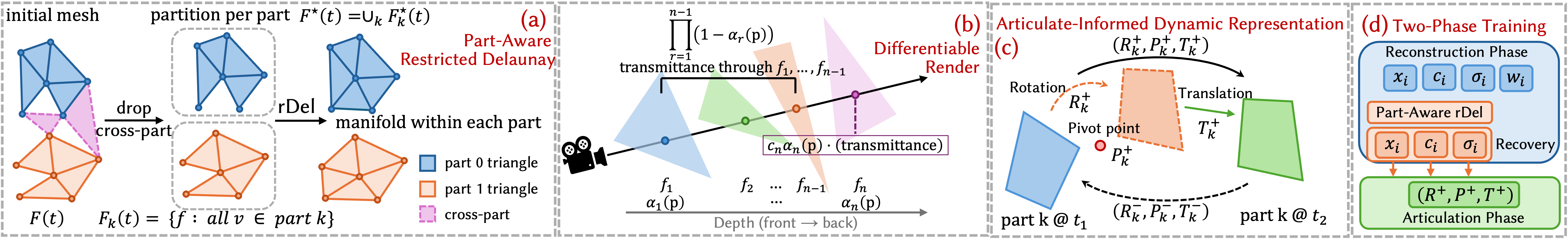}
   \caption{\textbf{Method sub-components.}
    \textbf{(a) Part-Aware Restricted Delaunay:} cross-part triangles (purple, dashed) are dropped and restricted Delaunay runs per cluster, yielding $F^{\star}(t)$, manifold within each part.
    \textbf{(b) Differentiable Render:} the mesh faces $f_1,\dots,f_n$ overlapping a pixel $\mathbf{p}$ are ordered front-to-back by depth. Each face contributes its color-opacity product $c_n\alpha_n(\mathbf{p})$, attenuated by the transmittance $\prod_{r<n}(1-\alpha_r(\mathbf{p}))$ of the faces in front. These contributions are alpha-composited into the pixel color (Eqs.~\eqref{eq:face_alpha},~\eqref{eq:alpha_compositing}).
    \textbf{(c) Articulation Dynamic Field:} the forward rigid transform $(R_k^{+},P_k^{+},T_k^{+})$ carries part $k$ from state $t_1$ (blue) to state $t_2$ (green): a rotation $R_k^{+}$ about pivot $P_k^{+}$, then a translation $T_k^{+}$. The backward transform $(R_k^{-},P_k^{-},T_k^{-})$ (dashed) is its analytic inverse, mapping $t_2$ back to $t_1$ (Eqs.~\eqref{eq:forward},~\eqref{eq:inverse_affine_motion}).
    \textbf{(d) Two-Phase Training:} the \emph{reconstruction phase} optimizes mesh attributes, then runs Part-Aware rDel; the \emph{articulation phase} freezes geometry and optimizes only $(R^{+}, P^{+}, T^{+})$.}
    \label{fig:render}
\end{figure*}

\mM\textbf{Part-Aware Articulated Mesh Field.} We reconstruct a part-aware mesh field $\mathcal{M}(t)=\left(V(t), F(t)\right)$ for each state $t\in\{t_1,t_2\}$, where the face set $F(t)\subset V(t)\times V(t)\times V(t)$ specifies the triangle connectivity over the vertices. Each vertex $v_i(t)=\left(x_i(t), c_i(t), \sigma_i(t), s_i(t)\right)\in V(t)$ stores a 3D position $x_i(t)\in\mathbb{R}^3$, spherical-harmonic color coefficients $c_i(t)$, opacity $\sigma_i(t)\in[0,1]$, and part logits $s_i(t)\in\mathbb{R}^{K+1}$ over $K+1$ rigid components, including one static base part and $K$ movable parts.
The soft part weights $w_i(t)=\operatorname{softmax}\!\left(s_i(t)\right)$ are hardened after the reconstruction stage to labels $p_i(t)=\arg\max_{k\in\{0,\ldots,K\}} w_{i,k}(t)$, which partition the vertices into part-specific sets $V_k(t)=\left\{v_i(t)\in V(t) : p_i(t)=k\right\}$ with $V(t)=\bigcup_{k=0}^{K} V_k(t)$. The opacity $\sigma_i(t)$ and spherical-harmonic color $c_i(t)$ are the vertex attributes alpha-composited by the differentiable mesh renderer used for the image losses (Fig.~\ref{fig:render}(b); Eqs.~\eqref{eq:face_alpha},~\eqref{eq:alpha_compositing}).
The articulation field (Fig.\ref{fig:render} (c)) assigns each part $k\in\{0,\dots,K\}$ a forward rigid motion consisting of a rotation $R_k^{+}\in\mathrm{SO}(3)$, a pivot $P_k^{+}\in\mathbb{R}^{3}$, and a translation $T_k^{+}\in\mathbb{R}^{3}$. Given the hardened part assignment $p(v_i)=\arg\max_{k} w_{i,k}$, each vertex is mapped from state $t_1$ to state $t_2$ by the rigid transform of its assigned part (Eq.~\eqref{eq:forward}). For revolute joints, a non-identity $R_k^{+}$ with $T_k^{+}=0$ encodes a pure hinge; for prismatic joints, $R_k^{+}=I$ and $T_k^{+}$ encodes a slider.

\mM\textbf{Part-aware Restricted Delaunay Remeshing.}
Global remeshing is suitable for static scenes  but unsafe for articulated objects. Triangles may connect vertices from different parts and shear when those parts move independently.
To make the topology compatible with articulation, we adapt restricted Delaunay remeshing~\cite{delaunay} to the articulated setting by running it independently within each articulated part, eliminating cross-part triangles that deform incorrectly under rigid motion.

Given the hardened part labels, we first split the face set by part:
\begin{equation}
F_k(t)
=
\left\{
f=(i,j,l)\in F(t)
:
p_i(t)=p_j(t)=p_l(t)=k
\right\}.
\label{eq:part_face_set}
\end{equation}
Faces whose vertices belong to different parts are discarded. We then run restricted Delaunay remeshing separately on each part:
\begin{equation}
F_k^{\star}(t)
=
\operatorname{rDel}
\left(
V_k(t), F_k(t)
\right),
\qquad
k=0,\ldots,K.
\label{eq:part_aware_rdel}
\end{equation}
The final topology is the union of all remeshed part topologies, $F^{\star}(t)=\bigcup_{k=0}^{K} F_k^{\star}(t)$, giving $\mathcal{M}^{\star}(t)=\left(V(t), F^{\star}(t)\right)$.
This operation changes only connectivity. Vertex positions, colors, opacities, and part labels remain attached to their original vertices.
As a result, $\mathcal{M}^{\star}(t)$ is composed of connected per-part submeshes.

\subsection{Articulation-aware Motion Consistency Learning}
\label{method:consistency}
The attributes attached to vertices in $\mathcal{M}(t_1)$ and $\mathcal{M}(t_2)$ are supervised by their own state-specific observations, but moving-part geometry and appearance may drift across states. Therefore, optimizing the articulation only along one direction does not guarantee that the implied inverse motion also explains the opposite-state observations.
\mC To tackle it, we introduce a bidirectional motion-consistency learning (Fig.~\ref{fig:arch}(c)) with:
\textbf{vertex-wise motion consistency}, directly constrains the transported mesh vertices in 3D;
and \textbf{pixel-wise motion consistency}, constrains the same part-wise motion through differentiable rendering. Both share the same forward rigid transforms and analytic inverse transforms, preventing the forward and backward directions from drifting.

\subsubsection{Vertex-wise Motion Consistency}
 We transport each vertex by the rigid motion of its assigned part and require it to be consistent with the corresponding surface in the other articulation state.

\textbf{Articulation Dynamic Field.} Let $p_i \in \{0,\ldots,K\}$ denote the hardened part label of vertex $v_i$. For each part $k$, we write the forward motion from state $t_1$ to state $t_2$ as an affine rigid transform (Fig.~\ref{fig:render}(c))

\begin{equation}
g_k^{+}(x) \;=\; R_{k}^{+}\bigl(x - P_{k}^{+}\bigr) + P_{k}^{+} + T_{k}^{+}=R_{k}^{+}x+b_k^{+}.
\label{eq:forward}
\end{equation}
where $b_k^{+}=-R_{k}^{+}P_{k}^{+}+P_{k}^{+}+T_{k}^{+}$.
Thus, the forward transported position of vertex $v_i \in V_{p_i}(t_1)$ is
\begin{equation}
\hat{x}_i^{\,1\rightarrow2}
=
g_{p_i}^+\!\left(x_i(t_1)\right)
=
R_{p_i}^+x_i(t_1)+b_{p_i}^+ .
\label{eq:forward_vertex_transport}
\end{equation}

The backward motion $(R_k^{-},T_k^{-},P_k^{-})$ that maps state $t_2$ back to state $t_1$ is not independently learned: it is the analytic inverse of the forward affine transform, $g_k^-(y)=(R_k^+)^\top(y-b_k^+)=R_k^-y+b_k^-$, with $R_k^-=(R_k^+)^\top$, $b_k^-=-(R_k^+)^\top b_k^+$, and, in pivot form with $P_k^-=P_k^+$, $T_k^-=-(R_k^+)^\top T_k^+$ (full homogeneous-coordinate derivation in Sec.~\ref{app:method-details}).

\textbf{Bidirectional Vertex Consistency.} For each transported vertex $\hat{x}_i^{\,1\rightarrow2}$, we search only within the same part $p_i$. Let
\begin{equation}
f_i^{\,1\rightarrow2}
=
\arg\min_{f\in F_{p_i}^{\star}(t_2)}
d\!\left(\hat{x}_i^{\,1\rightarrow2}, f\right)
\label{eq:closest_target_triangle_forward}
\end{equation}
be the closest triangle in the target-state submesh of the same part, and let $\beta_{ij}^{\,1\rightarrow2}$ be the barycentric weights of the closest point on this triangle. We interpolate the target vertex attributes (Fig.~\ref{fig:arch}(c)) as
\begin{equation}
\begin{aligned}
\tilde{a}_i^{\,1\rightarrow2}
&=
\sum_{j\in f_i^{\,1\rightarrow2}}
\beta_{ij}^{\,1\rightarrow2}a_j(t_2),
\qquad
a_j(t) = \left(x_j(t),c_j(t),\sigma_j(t)\right),\\
\tilde{a}_i^{\,1\rightarrow2}&=\left(\tilde{x}_i^{\,1\rightarrow2},\tilde{c}_i^{\,1\rightarrow2},\tilde{\sigma}_i^{\,1\rightarrow2}\right).
\end{aligned}
\label{eq:target_attribute_interpolation_forward}
\end{equation}

The forward vertex-wise motion-consistency loss is then
{\small
\begin{equation}
\begin{aligned}
\mathcal{L}_{\mathrm{vtx}}^{1\rightarrow2}
=
&\frac{1}{|V(t_1)|}
\sum_{v_i\in V(t_1)}
\eta_i^{\,1\rightarrow2}
\Big[
\lambda_x
\left\|
\hat{x}_i^{\,1\rightarrow2}
-
\tilde{x}_i^{\,1\rightarrow2}
\right\|_2^2\\
&\qquad
+
\lambda_c
\left\|
c_i(t_1)
-
\tilde{c}_i^{\,1\rightarrow2}
\right\|_2^2
+
\lambda_\sigma
\left|
\sigma_i(t_1)
-
\tilde{\sigma}_i^{\,1\rightarrow2}
\right|^2
\Big],
\end{aligned}
\label{eq:vertex_motion_consistency_forward}
\end{equation}
}
\noindent where $\eta_i^{\,1\rightarrow2}$ is a validity indicator that masks vertices without a valid same-part target correspondence, the geometric term pulls each transported vertex $\hat{x}_i^{\,1\rightarrow2}$ toward its interpolated target position $\tilde{x}_i^{\,1\rightarrow2}$, and the color and opacity terms enforce appearance consistency under the rigid part motion.
The backward direction is defined symmetrically: each vertex $v_j\in V(t_2)$ is transported back to state $t_1$ using the inverse motion, $\hat{x}_j^{\,2\rightarrow1}=g_{p_j}^-\!\left(x_j(t_2)\right)$.
We then find the closest same-part triangle in $F_{p_j}^{\star}(t_1)$, interpolate the corresponding target attributes, and compute $\mathcal{L}_{\mathrm{vtx}}^{2\rightarrow1}$ similarly to Eq.~\ref{eq:vertex_motion_consistency_forward}. This term directly supervises the articulated motion in 3D triangle vertex space.

\subsubsection{Pixel-wise Motion Consistency}
The pixel-wise motion-consist\-ency supervises the same part-wise affine motion with differentiable rendering. With the renderer $\mathcal{R}$, applying the forward part motions to the state $t_1$ mesh gives the articulated source mesh
\begin{equation}
\begin{aligned}
\hat{\mathcal{M}}^{1\rightarrow2}
&=
G^+\!\left(\mathcal{M}(t_1)\right)\\
&=
\left(
\left\{
\hat{x}_i^{\,1\rightarrow2},
c_i(t_1),
\sigma_i(t_1)
\right\}_{v_i\in V(t_1)},
F^\star(t_1)
\right),
\end{aligned}
\label{eq:forward_articulated_mesh}
\end{equation}
where each vertex position is transformed by the affine motion of its assigned part as in Eq.~\eqref{eq:forward_vertex_transport}. For a camera $\Pi_m^{t_2}$ observing state $t_2$, we render the RGB image $\hat{I}_m^{\,1\rightarrow2}=\mathcal{R}_{I}\!\left(\hat{\mathcal{M}}^{1\rightarrow2},\Pi_m^{t_2}\right)$, where $\mathcal{R}_{I}$ denotes the RGB channel of the renderer $\mathcal{R}$ in Eq.~\eqref{eq:mesh_renderer_outputs}. The forward pixel-wise motion-consistency loss compares the rendered observations with the true ones at state $t_2$:
\begin{equation}
\mathcal{L}_{\mathrm{pix}}^{1\rightarrow2}
=
\frac{1}{|\mathcal{C}_{t_2}|}
\sum_{m\in\mathcal{C}_{t_2}}
\ell_{\mathrm{pix}}
\left(
\hat{I}_m^{\,1\rightarrow2},
I_m(t_2)
\right),
\label{eq:pixel_motion_consistency_forward}
\end{equation}
where the per-view loss $\ell_{\mathrm{pix}}$ combines an $L_1$ RGB term and an SSIM term,
\begin{equation}
\ell_{\mathrm{pix}}(\hat I, I)=\lambda_{\mathrm{rgb}}\left\|\hat I-I\right\|_1+\lambda_{\mathrm{ssim}}\left(1-\mathrm{SSIM}(\hat I,I)\right).
\label{eq:pixel_motion_loss}
\end{equation}
The backward pixel-wise motion-consistency term is defined analogously. The inverse motion $G^-$ is applied to the state-$t_2$ mesh to produce $\hat{\mathcal{M}}^{2\rightarrow1}$, which is then rendered from state-$t_1$ cameras $\Pi_m^{t_1}$ and compared against the true state-$t_1$ observations $I_m(t_1)$ to form $\mathcal{L}_{\mathrm{pix}}^{2\rightarrow1}$, with the same per-view loss $\ell_{\mathrm{pix}}$ as in Eq.~\eqref{eq:pixel_motion_consistency_forward}.

This loss provides dense pixel-level supervision for the same part-wise affine motion used in the vertex-wise term.
The final motion-consistency objective sums the two vertex-wise and two pixel-wise terms,
\begin{equation}
\mathcal{L}_{\mathrm{motion}}
=\lambda_{\mathrm{VMC}}\left(\mathcal{L}_{\mathrm{vtx}}^{1\rightarrow2}+\mathcal{L}_{\mathrm{vtx}}^{2\rightarrow1}\right)
+\lambda_{\mathrm{PMC}}\left(\mathcal{L}_{\mathrm{pix}}^{1\rightarrow2}+\mathcal{L}_{\mathrm{pix}}^{2\rightarrow1}\right).
\label{eq:motion_total}
\end{equation}

The two phases (Fig.~\ref{fig:render}(d)) form one end-to-end training run, with the articulation-phase freeze preventing articulation signals from corrupting canonical geometry that has already converged; exact iteration budgets are in Sec.~\ref{app:training}.
\subsection{Differentiable Mesh Rendering and Optimization}
For a camera $\Pi_m^t$ at state $t$, mesh renderer $\mathcal{R}$ produces RGB, depth, and part-mask predictions:
\begin{equation}
\left(
\hat{I}_m(t), \hat{D}_m(t), \hat{S}_m(t)
\right)
=
\mathcal{R}
\left(
\mathcal{M}^{\star}(t), \Pi_m^t
\right).
\label{eq:mesh_renderer_outputs}
\end{equation}
For rendering, the faces overlapping a pixel are ordered front-to-back and alpha-composited with opacity-weighted smooth triangle window functions (Fig.~\ref{fig:render}(b)), detailed in Sec.~\ref{app:method-details}.
We optimize each state's mesh with a differentiable mesh renderer $\mathcal{R}$ by minimizing a reconstruction loss between its rendered outputs and their observations (defined in Sec.~\ref{app:training}).
During training, for each camera $m \in \mathcal{C}_t$ at state $t$, we have the RGB image $I_m(t)$, depth map $D_m(t)$ observations, and semantic part map $S_m(t)$ (ground-truth or predicted by our PArtSAM).

\textbf{PArtSAM.} When part masks are not provided, building on PartSAM~\cite{partsam}, we propose a \textbf{P}art-level 3D-consistent \textbf{Art}iculation segmentation, \textbf{PArt}SAM, to predict them: it back-projects the posed RGB-D views of each state into a colored point cloud, uses the motion cues between the two states to separate movable from static surfaces and to split the movable region into individual parts, and renders the resulting point labels back into per-view part maps in the same format as $S_m(t)$. Full details are in Sec.~\ref{app:additional-exp}.

{In the reconstruction stage, we optimize vertex positions, radiance, opacity, and part logits for each state. We then harden part assignments, apply the per-part restricted Delaunay remeshing in Eq.~\eqref{eq:part_aware_rdel}, and run a short reconstruction stage with fixed topology to refine vertex positions and appearance after remeshing. In the articulation stage, the part-aware meshes $\mathcal{M}^{\star}(t_1)$ and $\mathcal{M}^{\star}(t_2)$ are fixed, and only the per-part rigid motion parameters (Fig.~\ref{fig:render} (c)) are optimized using the vertex-wise and pixel-wise motion-consistency losses defined in Sec.~\ref{method:consistency}. Geometry is frozen here, so we supervise with RGB and the vertex-wise consistency only. The reconstruction objective, mesh initialization, and hyperparameters are in Sec.~\ref{app:training}.}

\section{Experiments}
\label{sec:exp}

\begin{figure}[t]
    \centering
    \includegraphics[width=\linewidth]{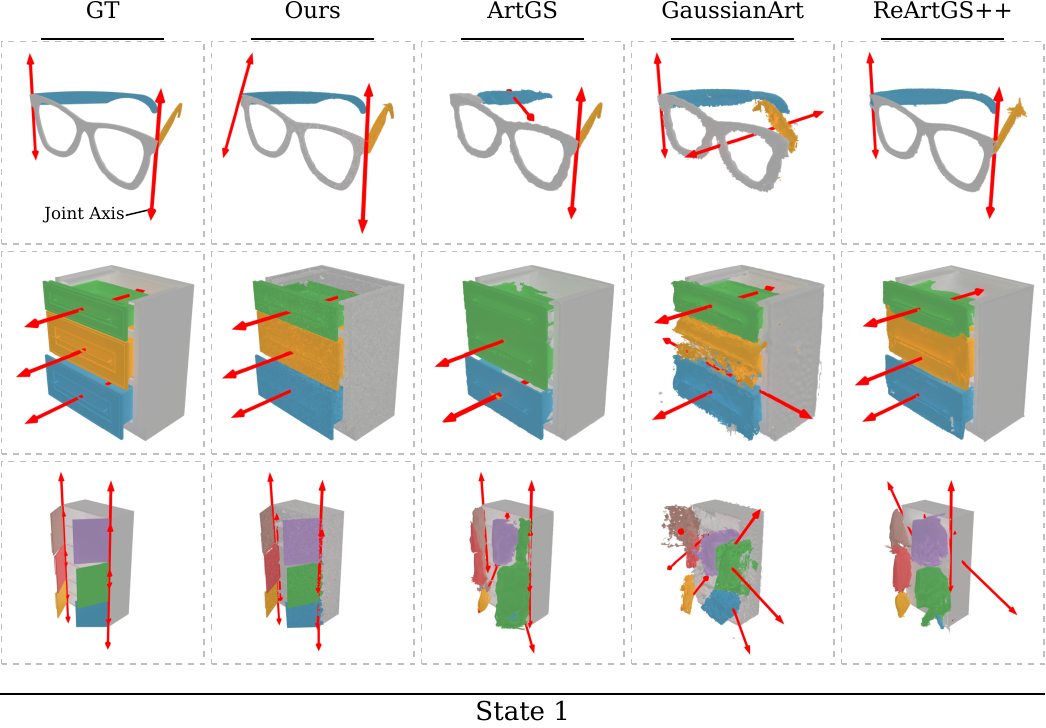}
    \caption{Qualitative comparisons on multi-part Articulate-100 objects (state-1 meshes); both states and more samples in Fig.~\ref{fig:articulate100-qual-full}. Red arrows mark predicted joint axes (lengths not meaningful). All methods use ground-truth part masks.}
    \label{fig:articulate100-qual}
\end{figure}

\subsection{Benchmarks and Implementation Details}
\label{sec:exp:setup}

\mB\textbf{Articulate-100.}
\label{sec:articulate100}
We construct \textbf{Articulate-100} (Fig.~\ref{fig:benchmark} and Sec.~\ref{app:articulate100}), a benchmark of 100 articulated objects sampled from PartNet-Mobility~\cite{sapien} across 16 categories, from single-joint objects (\emph{Laptop}, \emph{Knife}, \emph{Pliers}) to multi-part furniture. The distribution is dominated by \emph{Table} (34) and \emph{StorageFurniture} (42), to expose how each method scales with articulation complexity; Fig.~\ref{fig:benchmark} shows sample objects and the category and part-count distributions.

\textbf{PARIS.}
For comparability with prior work and real-world evaluation, we additionally use the PARIS benchmark~\cite{paris}, containing 10 synthetic and 2 real articulated objects with two-part articulation (one revolute or prismatic joint).

\textbf{Baselines.}
On Articulate-100, we compare \ArtMesh{} against three articulated 3DGS pipelines: \textbf{ArtGS}~\cite{artgs}, \textbf{GaussianArt}~\cite{gaussianart}, and \textbf{ReArtGS++}~\cite{reartgspp}.
\ArtMesh{} and GaussianArt consume part masks, which can be ground-truth masks or masks predicted by a segmentation frontend (our finetuned PArtSAM; Sec.~\ref{app:additional-exp}). ArtGS and ReArtGS++ take no part masks; we initialize their part centers from ground-truth masks and on Articulate-100 also report them as released.

\textbf{Metrics.}
We report five metrics: \textbf{Axis Ang} ($^\circ$), angular deviation between predicted and ground-truth joint axes; \textbf{Axis Pos} (0.1\,m), distance between the predicted and ground-truth axis lines (revolute joints only); \textbf{Part Motion} ($^\circ$ for revolute, $m$ for prismatic), joint-state error; \textbf{CD-s} (mm), Chamfer Distance on static parts; and \textbf{CD-m} (mm), Chamfer Distance on movable parts. Chamfer Distances are computed on 10{,}000 points sampled from predicted and ground-truth meshes.

\subsection{Results on Articulate-100}
\label{sec:exp:articulate100}

\definecolor{lightblue}{RGB}{225, 232, 240}

\begin{table*}[t]
\centering
\caption{Quantitative results on Articulate-100. Left: methods given GT part masks (GaussianArt, and ArtGS / ReArtGS++ with their part centers initialized from GT masks). Right: methods without GT masks (Ours-PArtSAM and GaussianArt-PArtSAM use PArtSAM masks; ArtGS / ReArtGS++ as released). All methods use the same matching: GT and predicted movable parts are paired one-to-one by Hungarian matching on CD-m, and the pairing is reused for every metric; an unmatched GT part receives the failure penalty. \textbf{Best} per column within each half in bold. Joints are scored as predicted (ArtGS and Ours: their own predicted types; GaussianArt, which only fixes a part's rotation to identity when it stays small, and ReArtGS++, which outputs a general rigid motion, carry no explicit type and are read as revolute if rotating ${\geq}5^\circ$); a wrong type takes the Part Motion and Axis Pos failure penalties. Gain is the improvement of Ours over the best baseline in the same half.}
\label{tab:articulate100}
\setlength{\tabcolsep}{3pt}
\renewcommand{\arraystretch}{0.85}
\resizebox{\linewidth}{!}{
\begin{tabular}{ll rrrrr @{\hskip 10pt} l rrrrr}
\toprule
 & \multicolumn{6}{c}{\textbf{With GT masks}} & \multicolumn{6}{c}{\textbf{Without GT masks}} \\
\cmidrule(lr){2-7}\cmidrule(lr){8-13}
 & Method & 2 (34) & 3 (20) & 4-5 (31) & 6+ (15) & All (100) & Method & 2 (34) & 3 (20) & 4-5 (31) & 6+ (15) & All (100) \\
\midrule
Axis Ang~($^\circ$) $\downarrow$ & GaussianArt-GTMask & 52.52 & 38.21 & 28.05 & 33.05 & 39.15 & GaussianArt-PArtSAM & 52.28 & 58.43 & 38.72 & 37.21 & 47.04 \\
 & ArtGS-GTMask & 5.31 & 11.09 & 5.80 & 14.37 & 7.98 & ArtGS & 2.71 & 9.89 & \textbf{15.42} & 22.26 & \textbf{11.02} \\
 & ReArtGS++-GTMask & \textbf{0.13} & \textbf{3.81} & 10.04 & 22.02 & 7.22 & ReArtGS++ & \textbf{0.61} & \textbf{9.56} & 21.66 & 28.43 & 13.10 \\
 & \our{Ours-GTMask} & \our{2.02} & \our{4.30} & \our{\textbf{2.93}} & \our{\textbf{6.90}} & \our{\textbf{3.49}} & \our{Ours-PArtSAM} & \our{1.68} & \our{12.02} & \our{19.79} & \our{\textbf{20.97}} & \our{12.26} \\
\rowcolor{gray!12}  & Gain & $-$1.89 & $-$0.49 & +2.87 & +7.47 & +3.73 & Gain & $-$1.07 & $-$2.46 & $-$4.37 & +1.29 & $-$1.24 \\
\midrule
Axis Pos~(0.1\,m) $\downarrow$ & GaussianArt-GTMask & 5.88 & 2.90 & 1.42 & 0.60 & 3.11 & GaussianArt-PArtSAM & 5.88 & 5.50 & 3.41 & 3.33 & 4.66 \\
 & ArtGS-GTMask & 0.59 & 1.00 & 0.58 & 1.09 & 0.75 & ArtGS & 0.30 & 1.14 & 1.05 & 1.63 & 0.90 \\
 & ReArtGS++-GTMask & 0.07 & 0.80 & 0.34 & 0.77 & 0.40 & ReArtGS++ & 0.12 & 0.92 & 0.77 & 1.38 & 0.67 \\
 & \our{Ours-GTMask} & \our{\textbf{0.04}} & \our{\textbf{0.03}} & \our{\textbf{0.01}} & \our{\textbf{0.03}} & \our{\textbf{0.03}} & \our{Ours-PArtSAM} & \our{\textbf{0.06}} & \our{\textbf{0.34}} & \our{\textbf{0.46}} & \our{\textbf{0.72}} & \our{\textbf{0.34}} \\
\rowcolor{gray!12}  & Gain & +0.03 & +0.77 & +0.33 & +0.57 & +0.37 & Gain & +0.06 & +0.58 & +0.31 & +0.66 & +0.33 \\
\midrule
Part Motion~($^\circ$/m) $\downarrow$ & GaussianArt-GTMask & 52.94 & 27.50 & 12.18 & 13.63 & 29.32 & GaussianArt-PArtSAM & 52.94 & 49.57 & 30.79 & 30.06 & 41.97 \\
 & ArtGS-GTMask & 5.32 & 9.04 & 4.59 & 9.55 & 6.47 & ArtGS & 2.69 & 10.94 & 9.39 & 14.37 & 8.17 \\
 & ReArtGS++-GTMask & \textbf{4.72} & 7.74 & 2.83 & 7.97 & 5.23 & ReArtGS++ & 4.28 & 9.51 & 8.10 & 12.05 & 7.67 \\
 & \our{Ours-GTMask} & \our{5.82} & \our{\textbf{2.12}} & \our{\textbf{2.32}} & \our{\textbf{3.43}} & \our{\textbf{3.64}} & \our{Ours-PArtSAM} & \our{\textbf{1.26}} & \our{\textbf{3.60}} & \our{\textbf{6.68}} & \our{\textbf{8.79}} & \our{\textbf{4.54}} \\
\rowcolor{gray!12}  & Gain & $-$1.10 & +5.62 & +0.51 & +4.54 & +1.59 & Gain & +1.43 & +5.91 & +1.42 & +3.26 & +3.13 \\
\midrule
CD-s~(mm) $\downarrow$ & GaussianArt-GTMask & 26.82 & 23.84 & 25.66 & 27.46 & 25.96 & GaussianArt-PArtSAM & 25.11 & 26.52 & 33.78 & 29.69 & 28.77 \\
 & ArtGS-GTMask & 20.05 & 20.62 & 30.93 & 26.36 & 24.48 & ArtGS & 19.25 & 21.07 & \textbf{24.67} & 28.75 & 22.72 \\
 & ReArtGS++-GTMask & 22.82 & 24.62 & 31.81 & 31.86 & 27.32 & ReArtGS++ & 23.31 & 25.63 & 32.65 & 47.07 & 30.23 \\
 & \our{Ours-GTMask} & \our{\textbf{17.95}} & \our{\textbf{18.86}} & \our{\textbf{19.66}} & \our{\textbf{20.11}} & \our{\textbf{18.98}} & \our{Ours-PArtSAM} & \our{\textbf{18.60}} & \our{\textbf{19.62}} & \our{26.06} & \our{\textbf{21.79}} & \our{\textbf{21.60}} \\
\rowcolor{gray!12}  & Gain & +2.10 & +1.76 & +6.00 & +6.25 & +5.50 & Gain & +0.65 & +1.45 & $-$1.39 & +6.96 & +1.12 \\
\midrule
CD-m~(mm) $\downarrow$ & GaussianArt-GTMask & 67.26 & 57.10 & 50.50 & 50.09 & 57.45 & GaussianArt-PArtSAM & 70.99 & 103.63 & 89.13 & \textbf{84.69} & 85.19 \\
 & ArtGS-GTMask & 30.36 & 66.24 & 70.82 & 87.59 & 58.66 & ArtGS & 40.83 & 133.13 & 146.59 & 173.09 & 111.91 \\
 & ReArtGS++-GTMask & 34.87 & 45.31 & 72.90 & 109.78 & 59.98 & ReArtGS++ & 50.52 & 115.62 & 140.96 & 166.14 & 108.92 \\
 & \our{Ours-GTMask} & \our{\textbf{11.09}} & \our{\textbf{15.74}} & \our{\textbf{15.27}} & \our{\textbf{22.40}} & \our{\textbf{15.01}} & \our{Ours-PArtSAM} & \our{\textbf{15.13}} & \our{\textbf{79.87}} & \our{\textbf{85.67}} & \our{86.78} & \our{\textbf{60.69}} \\
\rowcolor{gray!12}  & Gain & +19.27 & +29.57 & +35.23 & +27.69 & +42.44 & Gain & +25.70 & +23.76 & +3.46 & $-$2.09 & +24.50 \\
\bottomrule
\end{tabular}
}
\end{table*}

\textbf{Qualitative results.}
Fig.~\ref{fig:articulate100-qual} and Fig.~\ref{fig:articulate100-qual-full} show qualitative comparisons on representative multi-part objects. Beyond three parts ArtGS can fracture movable meshes with misaligned joints on some objects, the axes of GaussianArt drift on objects with many similar parts, and \ArtMesh{} separates the parts cleanly with aligned axes.

\textbf{Quantitative results.}
\mC Table~\ref{tab:articulate100} reports per-bucket and aggregate results. With ground-truth masks, ReArtGS++ has lower Axis Ang and Part Motion on 2-part objects, while \ArtMesh{} has the lowest Axis Pos and far lower Chamfer Distances. With three or more parts \ArtMesh{} is best on every metric but 3-part Axis Ang, and best on all five in aggregate. ReArtGS++~\cite{reartgspp} lags on the remaining articulation metrics and on movable-part geometry, so its planar Gaussian articulation does not scale to many parts.
The failure patterns of ArtGS and GaussianArt match the limitations each paper identifies, amplified by our broader category mix; see Sec.~\ref{app:baseline-failure}.
Ours-PArtSAM uses masks predicted by our finetuned PArtSAM segmenter (Sec.~\ref{app:additional-exp}). It is on par with the ground-truth-mask rows on 2-part objects, and on multi-part objects mask noise raises the Chamfer Distances and Axis Ang (above ArtGS on 3 to 5 parts), while Part Motion stays below ArtGS.

\begin{figure}[!tb]
    \centering
    \includegraphics[width=\linewidth]{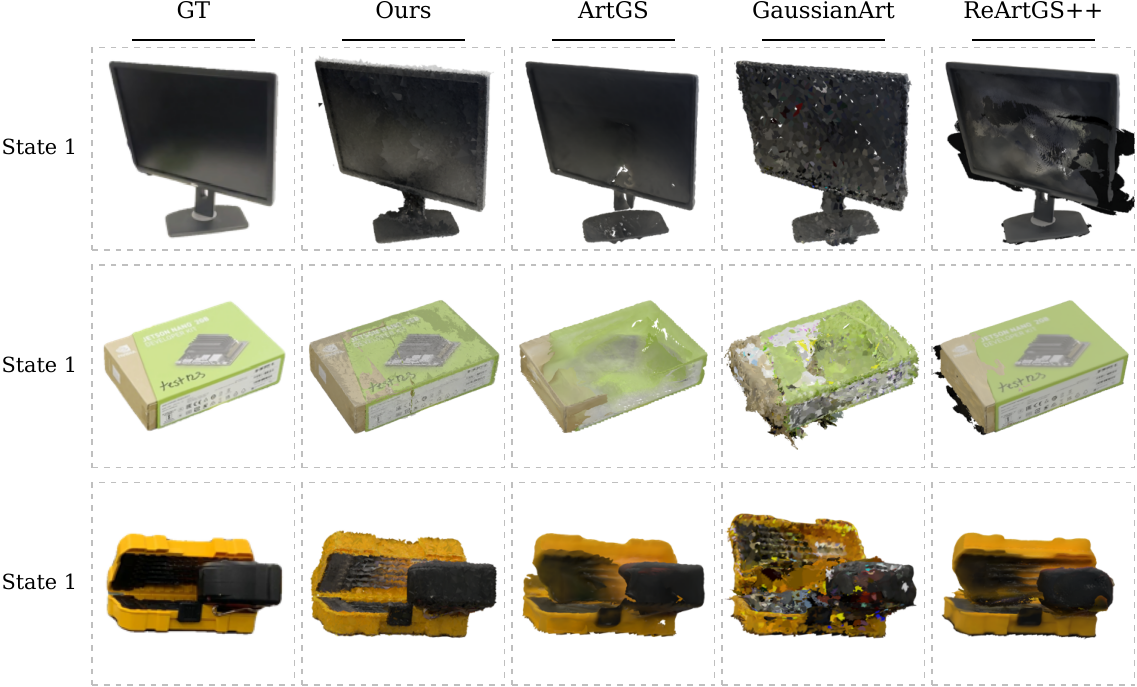}
    \caption{Real-world reconstructions on SplArt~\cite{splart} (end state; both states in Fig.~\ref{fig:splart}). \ArtMesh{} meshes are clean, while GaussianArt fragments surfaces, ReArtGS++ distorts the monitor, ArtGS washes out the package, and ArtGS and ReArtGS++ (given ground-truth part masks) blur the toolbox interior.}
\label{fig:splart_main}
\end{figure}

\textbf{Further experiments.} In Sec.~\ref{app:additional-exp}, estimated poses account for most of the error under COLMAP poses and estimated depth (Table~\ref{tab:robust-depthpose}), and \ArtMesh{} comes within 0.7\,dB of ArtGS's PSNR with a quarter of its CD-m (Table~\ref{tab:novel-view}) at ground-truth TLPIPS (Table~\ref{tab:tlpips}). On 9 real SplArt captures with hand-labelled joints, \ArtMesh{} recovers the correct joint type on every object and has the lowest mean Axis Ang, Axis Pos and Part Motion (Table~\ref{tab:splart_real}, Fig.~\ref{fig:splart_main}).

\subsection{Results on PARIS}
\label{sec:exp:paris}
\begin{figure}[t]
    \centering
    \includegraphics[width=\linewidth]{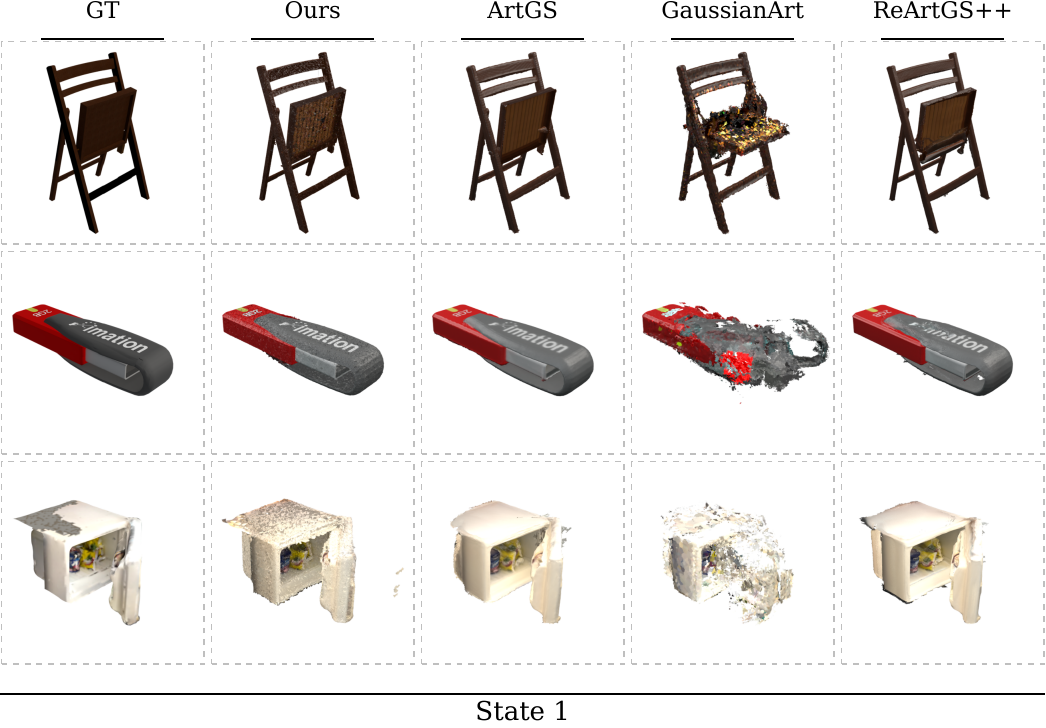}
    \caption{Results on PARIS~\cite{paris}, where \ArtMesh{}'s high-part-count advantages are least exercised. Meshes are rendered in Blender; all methods use ground-truth part masks.}
        \label{fig:paris}
\end{figure}
\definecolor{lightblue}{RGB}{225, 232, 240}
\begin{table}[t]
\centering
\caption{Paris benchmark results, averaged over all objects of each split with joint types weighted by object count (Syn: 8 revolute, 2 prismatic; Real: 1 each). All methods use ground-truth part masks (ArtGS and ReArtGS++ initialize their part centers from them). \textbf{Best}, \underline{Second}.}
\label{tab:paris}
\setlength{\tabcolsep}{3pt}
\renewcommand{\arraystretch}{1.05}
\resizebox{\linewidth}{!}{
\begin{tabular}{l rrrrr}
\toprule
 & Axis Ang\,$\downarrow$ & Axis Pos\,$\downarrow$ & Part Motion\,$\downarrow$ & CD-s\,$\downarrow$ & CD-m\,$\downarrow$ \\
\midrule
\textit{Syn (10)} &  &  &  &  &  \\
ArtGS       & \textbf{0.03} & \textbf{0.00} & \textbf{0.03} & \underline{20.62} & \underline{36.11} \\
ReArtGS++   & \underline{0.11} & 1.27 & 6.59 & 38.23 & 58.51 \\
GaussianArt & 70.39 & 8.00 & 72.00 & 24.18 & 59.29 \\
Ours        & \cellcolor{lightblue}0.68 & \cellcolor{lightblue}\underline{0.02} & \cellcolor{lightblue}\underline{0.54} & \cellcolor{lightblue}\textbf{19.59} & \cellcolor{lightblue}\textbf{12.44} \\
\midrule
\textit{Real (2)} &  &  &  &  &  \\
ArtGS       & \underline{2.82} & 0.23 & \underline{0.97} & \underline{27.36} & \textbf{50.38} \\
ReArtGS++   & 41.19 & \underline{0.20} & 1.66 & 41.51 & 52.91 \\
GaussianArt & 45.26 & 5.00 & 45.02 & 40.76 & 60.76 \\
Ours        & \cellcolor{lightblue}\textbf{2.37} & \cellcolor{lightblue}\textbf{0.07} & \cellcolor{lightblue}\textbf{0.66} & \cellcolor{lightblue}\textbf{27.25} & \cellcolor{lightblue}\underline{50.84} \\
\bottomrule
\end{tabular}
}
\end{table}

PARIS~\cite{paris} is a small two-part benchmark where the scaling and per-part isolation of \ArtMesh{} are least exercised (Table~\ref{tab:paris}, Fig.~\ref{fig:paris}, textured in Fig.~\ref{fig:paris-extra}). On the synthetic split, ArtGS has the lowest joint errors and \ArtMesh{} is close behind ($0.68^\circ$ Axis Ang, $0.54^\circ$ Part Motion); ReArtGS++ recovers the axis directions but has larger Axis Pos and Part Motion errors, and GaussianArt predicts no rotation on revolute joints, which are scored as type mismatches, the same cross-part blending failure observed on Articulate-100. \ArtMesh{} attains the best Chamfer Distances on both static and movable parts. On the real split, \ArtMesh{} is best on every metric except CD-m, where it trails ArtGS.

\subsection{Ablation Studies}
\label{sec:exp:ablation}
\begin{figure}[!tb]
    \centering
    \includegraphics[width=\linewidth]{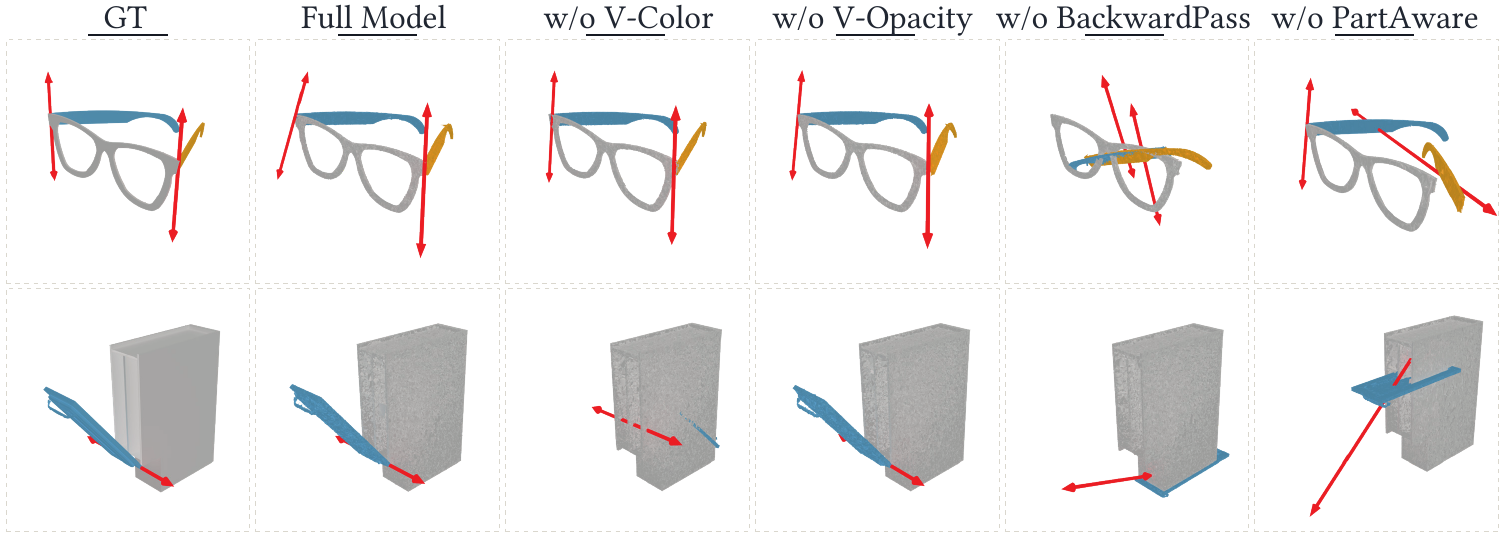}
    \caption{Ablation of four \ArtMesh{} components on Articulate-100: (i) \emph{color consistency}, (ii) \emph{opacity consistency}, (iii) the \emph{backward pass}, and (iv) \emph{part-aware restricted Delaunay}. See Table~\ref{tab:ablation}.}
        \label{fig:ablation}
\end{figure}
\begin{table}[!tb]
\centering

\caption{Ablation on Articulate-100. Best in \textbf{bold}.}
\label{tab:ablation}
\setlength{\tabcolsep}{4pt}
\renewcommand{\arraystretch}{1.1}
\resizebox{\linewidth}{!}{
\begin{tabular}{lccccc}
\toprule
 & Full Model
 & \shortstack{w/o\\[-2pt] V-Color}
 & \shortstack{w/o\\[-2pt] V-Opacity}
 & \shortstack{w/o\\[-2pt] Backward}
 & \shortstack{w/o\\[-2pt] Part-Aware} \\
\midrule
Axis Ang~($^\circ$) $\downarrow$     & \textbf{3.49}  & 6.02  & 6.87  & 10.79 & 7.10  \\
\midrule
Axis Pos~(0.1\,m) $\downarrow$     & \textbf{0.03}  & 0.09  & 0.18  & 0.48  & 0.29 \\
\midrule
Part Motion~($^\circ$/m) $\downarrow$  & \textbf{3.64}  & 5.61  & 5.88  & 7.36  & 6.10  \\
\midrule
CD-s~(mm) $\downarrow$         & 18.98          & 18.96 & 18.99 & 18.94 & \textbf{18.58} \\
\midrule
CD-m~(mm) $\downarrow$         & \textbf{15.01} & 21.91 & 19.86 & 29.06 & 19.30 \\
\bottomrule
\end{tabular}
}
\end{table}

\mM\mC We ablate four components (Table~\ref{tab:ablation}, Fig.~\ref{fig:ablation}). (1) \textbf{Vertex-wise appearance (color) consistency:} a key cross-state cue; removal raises Axis Ang and CD-m on weakly textured parts. (2) \textbf{Vertex-wise geometry (opacity) consistency:} removal degrades motion comparably with smaller geometry effects, mainly disambiguating which triangles move together. (3) \textbf{Backward pass:} the most damaging ablation (Axis Ang, Part Motion, CD-m); bidirectional supervision keeps the motion field self-consistent. (4) \textbf{Part-aware restricted Delaunay:} replacing it with global Delaunay degrades Axis Ang and CD-m and slightly improves CD-s, since its main role is to keep the consistency losses from pulling vertices across joints. Removing the recovery stage or the pixel-wise loss raises Axis Ang to $9.00^\circ$ and $46.64^\circ$, and Gaussian backbones raise it above $49^\circ$ with about twice the CD-m (Table~\ref{tab:finer-ablation}).

\subsection{Mesh Quality}
\label{sec:exp:meshquality}
\begin{table}[!tb]
\centering
\setlength{\tabcolsep}{4pt}
\caption{Self-Penetration Depth (SPD~$\downarrow$, fraction of the bounding-box diagonal). ArtGS, ReArtGS++, and GaussianArt have no direct mesh (N/A) and need TSDF extraction.}
\label{tab:spd}
\resizebox{0.9\linewidth}{!}{
\begin{tabular}{lcccc}
\toprule
 & ArtGS & GaussianArt & ReArtGS++ & Ours  \\
\midrule
Mesh $\downarrow$      & N/A    & N/A    & N/A & 0.0224 \\
TSDF $\downarrow$      & 0.0023 & 0.0055 & 0.0028 & \textbf{0.0018} \\
\bottomrule
\end{tabular}}
\end{table}

\mM Per-part connectivity gives the motion-consistency losses a rigid structure, so joints are estimated on the exported geometry, whereas ArtGS, ReArtGS++, and GaussianArt produce no mesh until post-hoc TSDF extraction. Our exported mesh needs no TSDF step (Fig.~\ref{fig:teaser}(d); it also exports as a URDF, Sec.~\ref{experiments:simulation}, and Fig.~\ref{fig:more-mesh-quality} shows the gaps TSDF fusion leaves on the baselines); for a like-for-like comparison we apply the same TSDF pass the baselines require, under which our mesh has the lowest Self-Penetration Depth (Table~\ref{tab:spd}).

A part hidden in both states can get a drifted axis or merge into its neighbor (Sec.~\ref{app:failure}).

\section{Conclusion}
\label{sec:conclusion}

We presented \ArtMesh{}, a method for reconstructing articulated objects as connected triangle meshes with per-part rigid motion from multi-view observations at two states. It rests on per-part remeshing, vertex-wise color and opacity consistency, and a bidirectional cycle. On Articulate-100 with ground-truth part masks, \ArtMesh{} has the best aggregate joint-axis and part-level geometry accuracy among the evaluated 3DGS baselines (ArtGS, ReArtGS++, and GaussianArt), with the largest gains on many-part objects. Accuracy drops with predicted masks, and the lowest self-penetration still needs a TSDF pass. Surface quality, segmentation reliance and multi-state sequences remain future work.

{
    \small
    \bibliographystyle{ieeenat_fullname}
    \bibliography{main}
}

\clearpage
\appendix
This appendix contrasts the surface representations (Sec.~\ref{app:gaussian-geo}), describes the Articulate-100 benchmark (Sec.~\ref{app:articulate100}) and the simulator export (Sec.~\ref{experiments:simulation}), reports additional experiments (Sec.~\ref{app:additional-exp}), analyzes baseline failures and our own (Secs.~\ref{app:baseline-failure} and~\ref{app:failure}), and lists training and method details (Secs.~\ref{app:training} and~\ref{app:method-details}).

\section{Gaussian- vs. Mesh-Splatting Geometry Representation}
\label{app:gaussian-geo}
\begin{figure*}[t]
    \centering
    \includegraphics[width=0.8\linewidth]{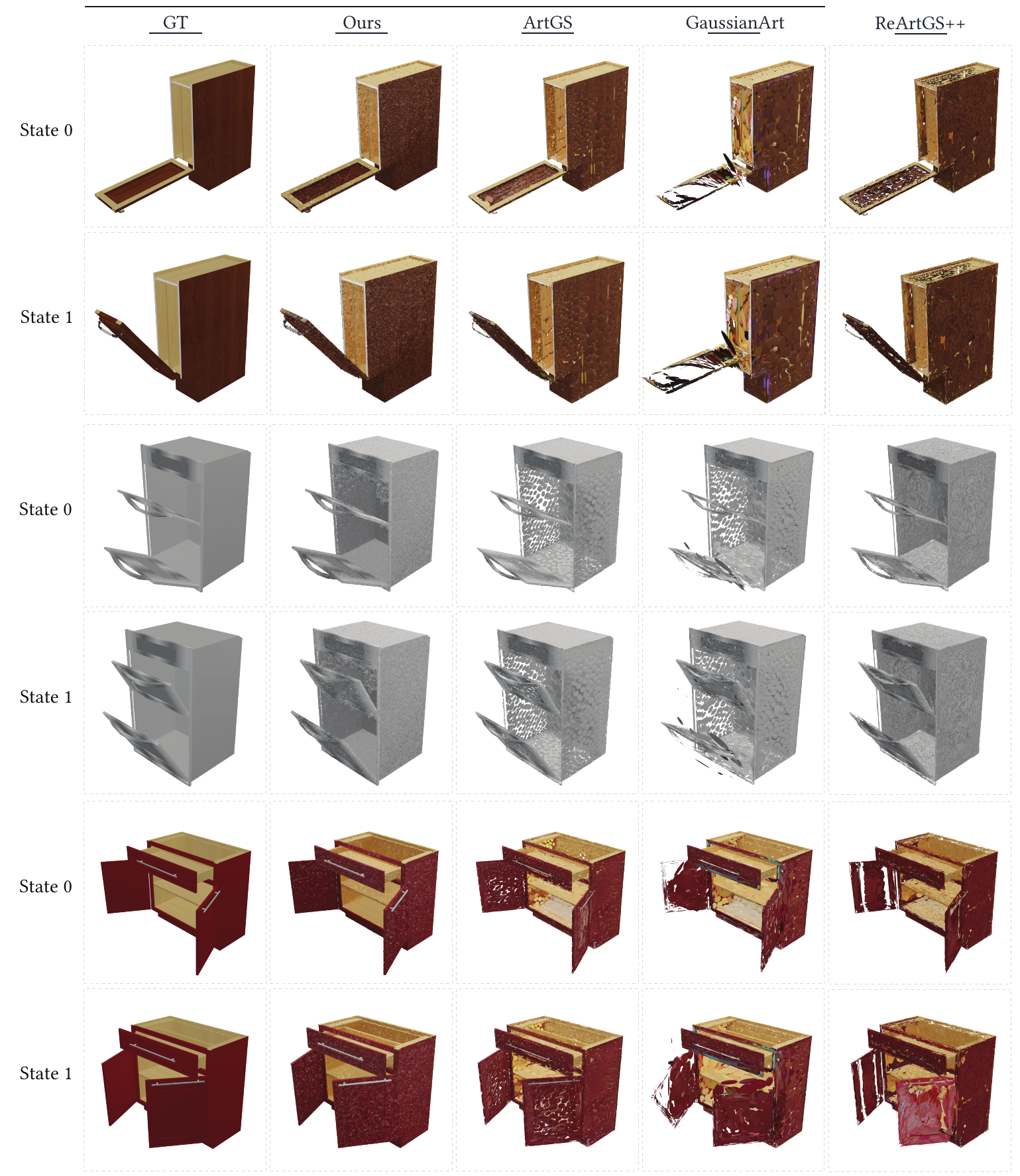}
    \caption{Qualitative comparison of reconstructed surfaces. ArtGS, ReArtGS++, and GaussianArt produce unstructured Gaussians that require post-hoc TSDF fusion, which introduces gaps, smoothing, and topological errors; \ArtMesh{} optimizes a connected, opaque triangle mesh end-to-end, so the trained surface is the exported surface.}
    \label{fig:more-mesh-quality}
\end{figure*}
Beyond joint and part-level metrics, the form of the reconstructed surface itself matters for downstream use. Figure~\ref{fig:more-mesh-quality} contrasts the surfaces produced by the four methods. The Gaussian-based pipelines, ArtGS~\cite{artgs}, ReArtGS++~\cite{reartgspp}, and GaussianArt~\cite{gaussianart}, optimize an unstructured cloud of disconnected primitives and rely on TSDF fusion of rendered depth maps to recover a triangle mesh after training; the resulting surfaces show visible inter-primitive gaps, density variation along part boundaries, and degraded geometry on thin or texture-less regions, with TSDF smoothing introducing additional discrepancies between what was optimized and what is exported. \ArtMesh{} resolves both issues by optimizing a connected mesh end-to-end: the surface used during training and the surface delivered to downstream tasks are identical.

\begin{figure*}[t]
    \centering
    \includegraphics[width=0.8\linewidth]{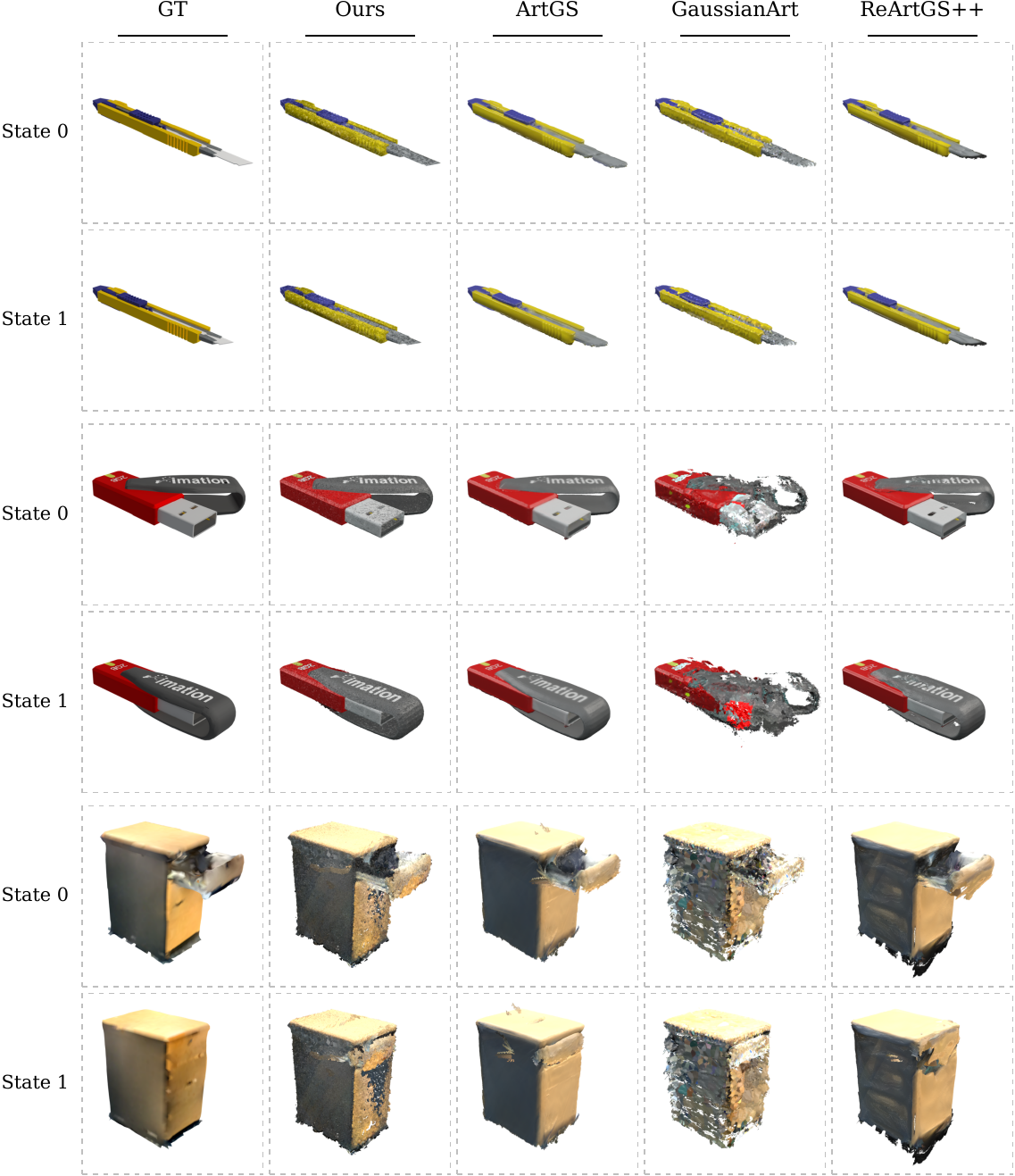}
    \caption{Additional textured qualitative comparisons between \ArtMesh, ArtGS~\cite{artgs}, GaussianArt~\cite{gaussianart}, and ReArtGS++ on the PARIS dataset, complementing Fig.~\ref{fig:paris}. All methods use ground-truth part masks.}
    \label{fig:paris-extra}
\end{figure*}

\section{Articulate-100 Benchmark}
\label{app:articulate100}

\begin{figure*}[t]
    \centering
    \includegraphics[width=\linewidth]{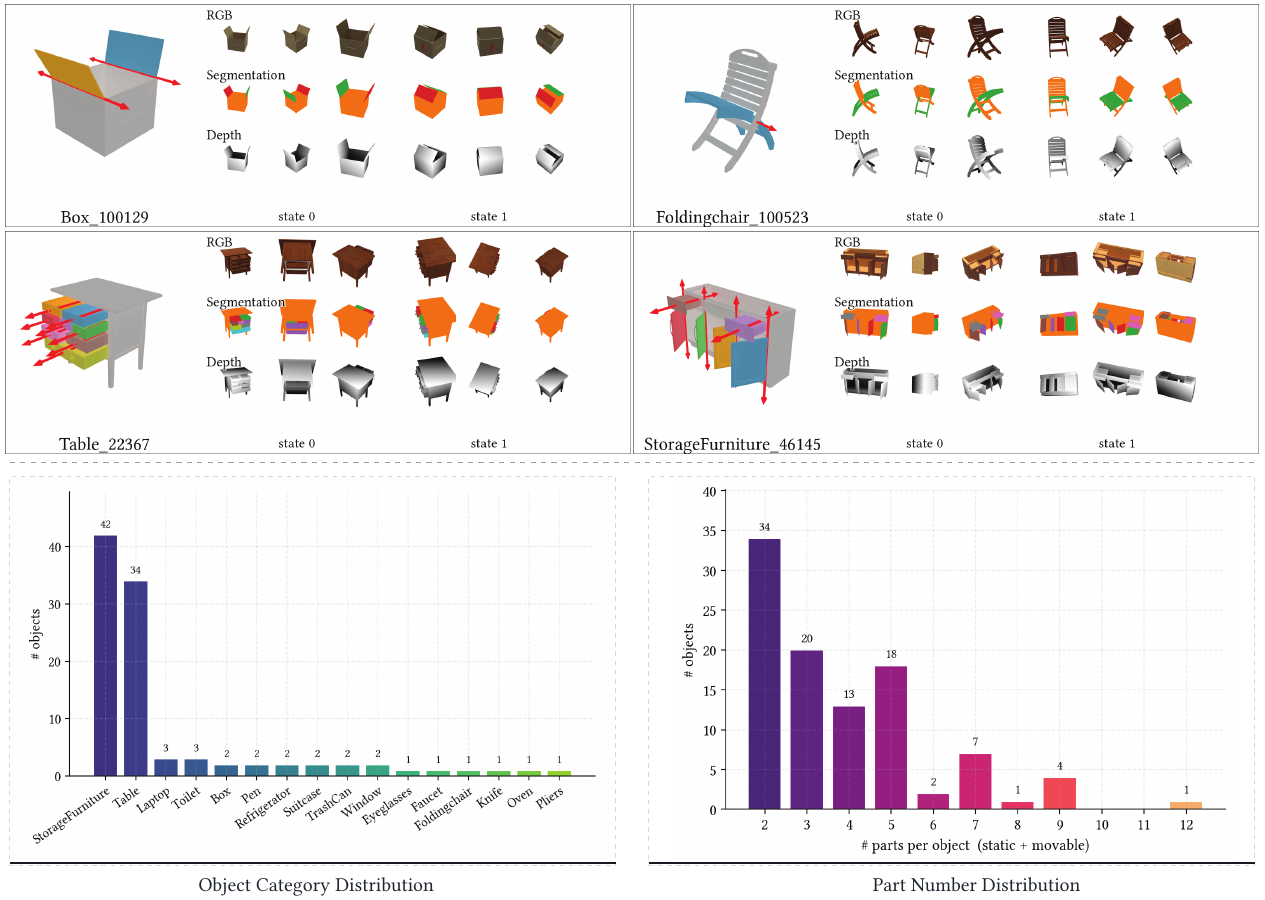}
    \caption{Benchmark overview. Each object provides RGB, depth, segmentation, and articulation annotations, alongside the dataset's part-count and category distributions.}
    \label{fig:benchmark}
\end{figure*}

Figure~\ref{fig:benchmark} shows sample objects from the dataset and presents the distribution of object categories and part numbers. Each object is rendered in Blender at two motion states, with the starting state sampled from $[0.60, 0.80]$ and the ending state from $[0.20, 0.40]$ on the normalized 1-DoF range (0 = fully closed, 1 = fully open). For each state we sample 100 training views and 20 test views on a spherical region around the object at $800\times800$ resolution, yielding posed RGB-D observations together with ground-truth motion parameters and part segmentation.
\begin{figure*}[t]
    \centering
    \includegraphics[width=0.8\linewidth]{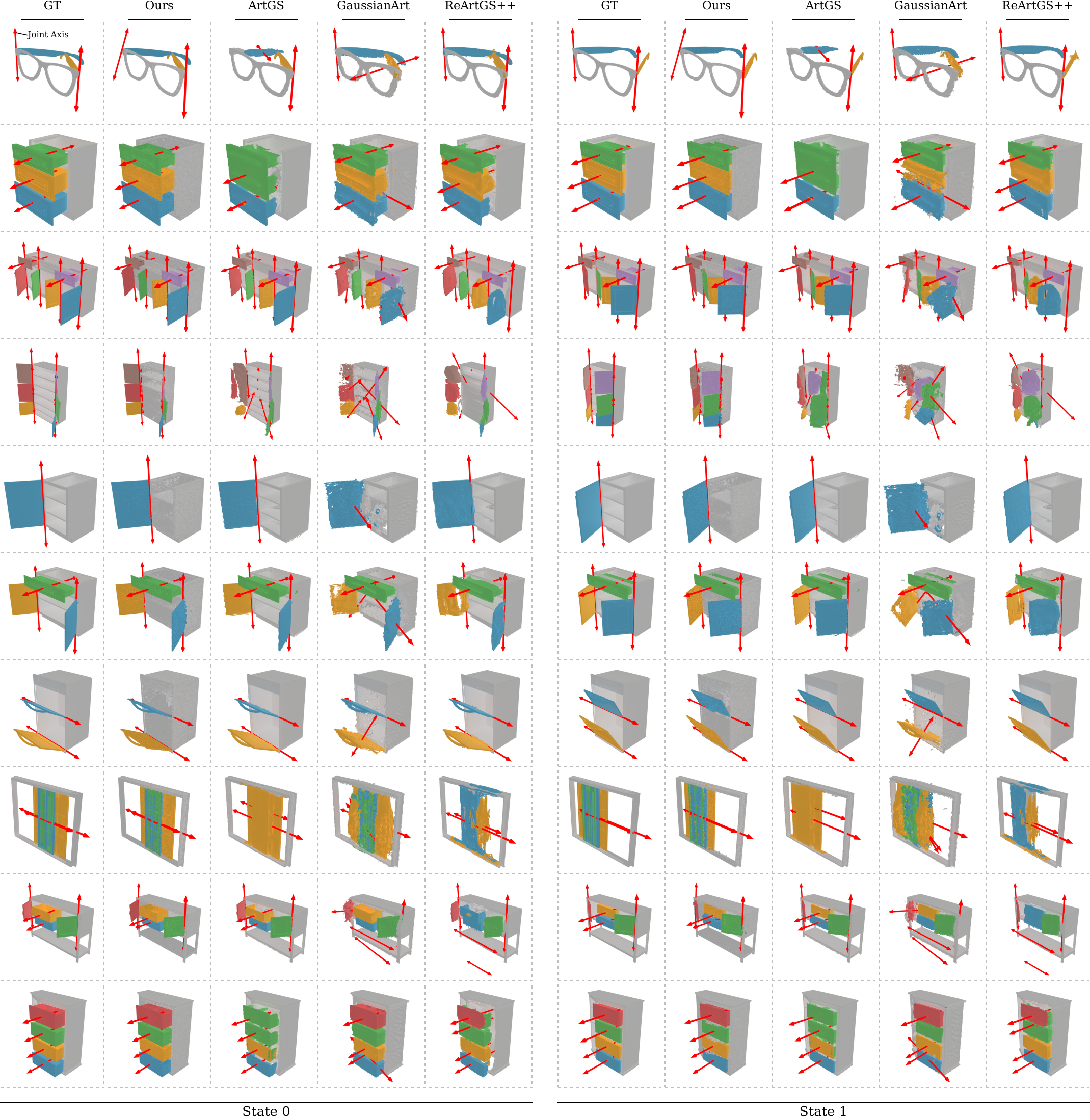}
    \caption{Full qualitative comparisons on Articulate-100 at both states, adding the state-0 counterparts of the state-1 meshes in Fig.~\ref{fig:articulate100-qual} and additional samples. All methods use ground-truth part masks.}
    \label{fig:articulate100-qual-full}
\end{figure*}

\section{Application in Simulation}
\label{experiments:simulation}

\begin{figure}[t]
    \centering
    \includegraphics[width=\linewidth]{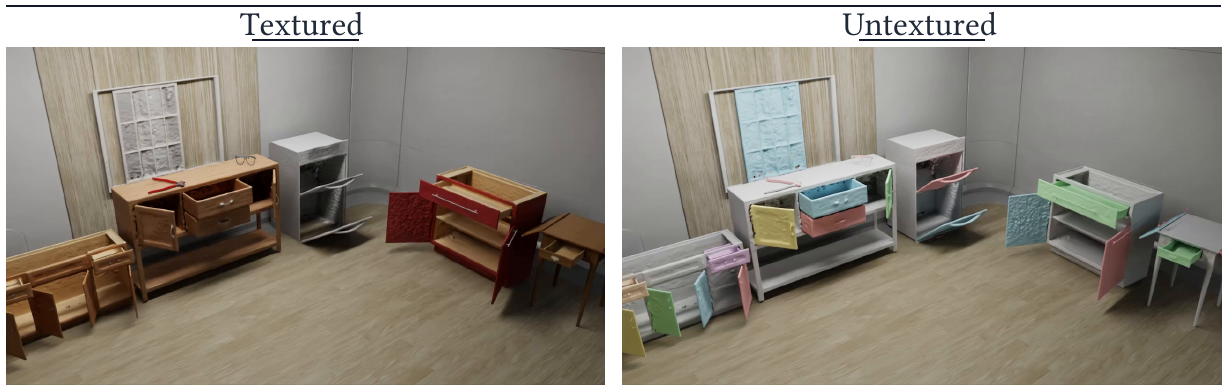}
    \caption{\ArtMesh{} reconstructions imported into NVIDIA Isaac Sim (TSDF meshes with a convex-hull collision shape per part).}
    \label{fig:simulation}
\end{figure}

From a trained \ArtMesh{} model, we export each per-part submesh as a Universal Robot Description Format (URDF) link and each $(R_k^{+},P_k^{+},T_k^{+})$ as a revolute or prismatic joint read directly from the optimized parameters. The URDF loads into NVIDIA Isaac Sim and supports physically plausible interaction (drawer pulls, door swings, lid openings; Fig.~\ref{fig:simulation}). For these scenes we use the TSDF meshes with a convex-hull collision shape per part; the directly exported mesh is evaluated in SAPIEN below. The Self-Penetration Depth reported in Table~\ref{tab:spd} of the main text is defined as $\mathrm{SPD}=\max_{s\in[0,1]}\max_{i\neq j} d_p\!\left(T_i(s)P_i,\,T_j(s)P_j\right)$, where $s$ moves between the two observed states, $P_i$ is part $i$'s mesh, $T_i(s)$ its rigid transform at state $s$, and $d_p$ the interpenetration depth; it is reported as a fraction of the object's bounding-box diagonal.

\section{Additional Experiments}
\label{app:additional-exp}

\textbf{Real-world validation.}
\begin{figure}[t]
    \centering
    \includegraphics[width=\linewidth]{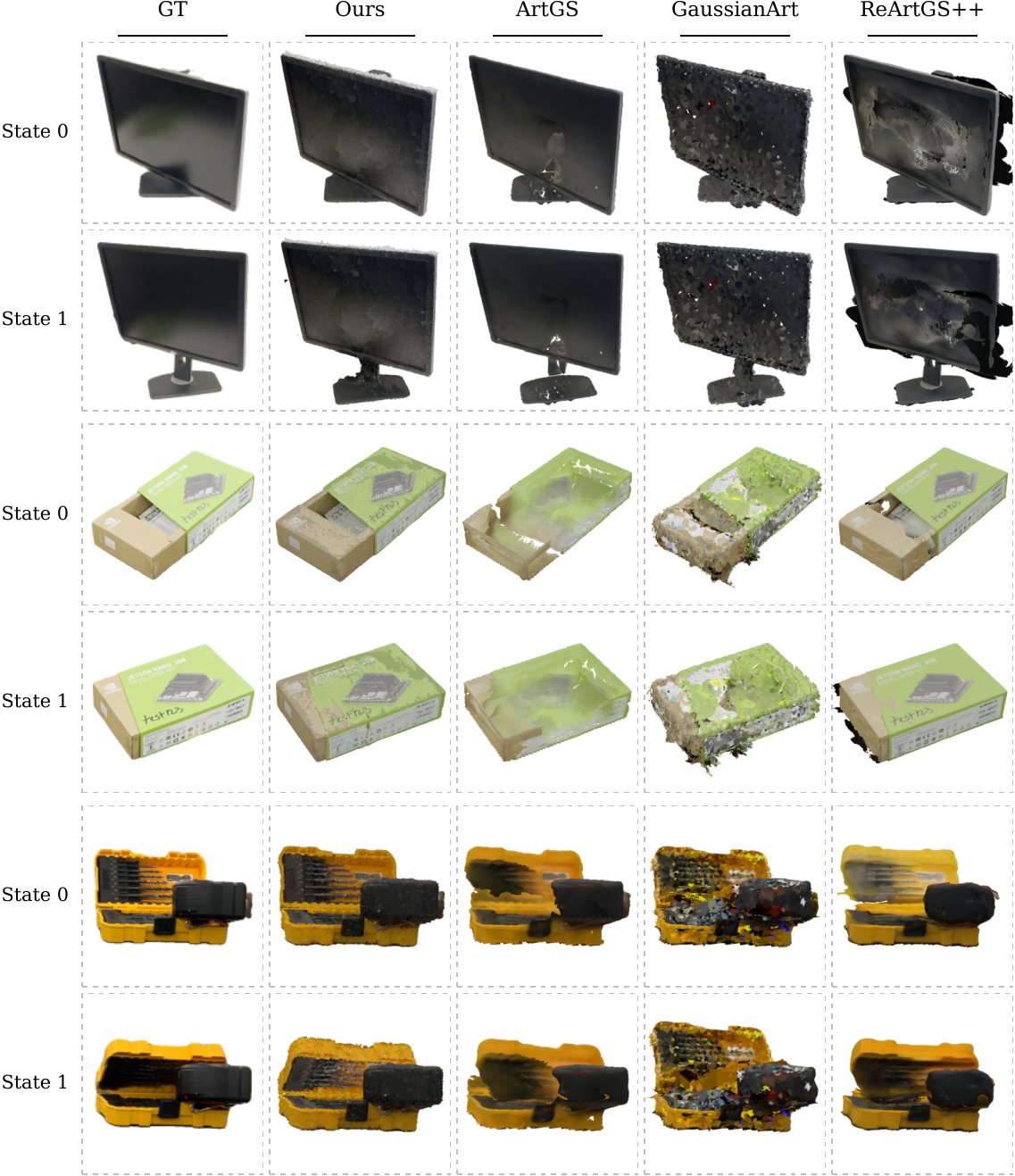}
    \caption{Real-world reconstructions on SplArt~\cite{splart} at both articulation states. \ArtMesh{} produces clean meshes from real captures, while GaussianArt yields fragmented, speckled surfaces, ReArtGS++ distorts the monitor screen, ArtGS washes out the package print, and ArtGS and ReArtGS++ blur the fine texture inside the toolbox. ArtGS and ReArtGS++ use ground-truth part masks.}
    \label{fig:splart}
\end{figure}
We further validate \ArtMesh{} on real captures from SplArt~\cite{splart}, which provides real captures of 7 objects with 9 articulations, each observed at two states with one moving joint, but no ground-truth meshes or joint parameters. Fig.~\ref{fig:splart} compares reconstructions at both states: \ArtMesh{} recovers clean, connected surfaces from real observations, while GaussianArt yields fragmented, speckled surfaces, ReArtGS++ distorts the monitor screen, ArtGS washes out the package print, and both ArtGS and ReArtGS++, here initialized from ground-truth part masks, blur the fine texture inside the toolbox. To evaluate the joints quantitatively, we label the joint of each of the 9 captures by triangulating points marked by hand on the hinge or the sliding part across calibrated views, and score every method with the metrics of Table~\ref{tab:articulate100} (Table~\ref{tab:splart_real}). \ArtMesh{} predicts the correct joint type on all 9 captures, as does ReArtGS++, while ArtGS mislabels the toolbox hinge as prismatic. \ArtMesh{} has the lowest mean Axis Ang ($5.22^\circ$ vs.\ $14.59^\circ$ for ArtGS and $15.46^\circ$ for ReArtGS++), Axis Pos and Part Motion.
\begin{table}[t]
\centering
\caption{Joint estimation on the 9 SplArt real captures (5 revolute, 4 prismatic), averaged over objects and scored against hand-triangulated joint labels with the metrics of Table~\ref{tab:articulate100}: Axis Ang ($^\circ$), Axis Pos (axis-line distance divided by the labelled hinge length, since the captures have no metric scale; prismatic joints count as 0) and Part Motion ($^\circ$ for revolute joints, scene units for prismatic joints). Chamfer Distances are not reported since SplArt has no ground-truth meshes. Joints are scored as predicted with the type rule of Table~\ref{tab:articulate100}; a wrong predicted type keeps its axis angle and takes the failure penalty on Axis Pos (1 hinge length) and Part Motion ($90^\circ$, or the labelled slide). ArtGS and ReArtGS++ are initialized from ground-truth part masks. \textbf{Best} in bold; Gain is the improvement of Ours over the best baseline.}
\label{tab:splart_real}
\setlength{\tabcolsep}{4pt}
\renewcommand{\arraystretch}{1.05}
\resizebox{0.85\linewidth}{!}{
\begin{tabular}{l rrrc}
\toprule
 & Axis Ang\,$\downarrow$ & Axis Pos\,$\downarrow$ & Part Motion\,$\downarrow$ & Correct type\,$\uparrow$ \\
\midrule
GaussianArt & 39.81 & 0.556 & 50.02 & 4/9 \\
ArtGS       & 14.59 & 0.169 & 13.04 & 8/9 \\
ReArtGS++   & 15.46 & 0.764 & 24.37 & \textbf{9/9} \\
\our{Ours}  & \our{\textbf{5.22}} & \our{\textbf{0.022}} & \our{\textbf{4.25}} & \our{\textbf{9/9}} \\
\rowcolor{gray!12} Gain & +9.37 & +0.147 & +8.79 & \\
\bottomrule
\end{tabular}
}
\end{table}

\textbf{Robustness to imperfect inputs.}
We evaluate robustness under predicted inputs: part masks predicted by PArtSAM, our articulation-adapted finetune of the promptable 3D part segmentation model PartSAM~\cite{partsam} (Ours-PArtSAM in Table~\ref{tab:articulate100}; pipeline detailed below), and COLMAP-estimated poses and Depth-Anything-V2 depth on the \emph{StorageFurniture} objects (Table~\ref{tab:robust-depthpose}). With predicted masks, errors grow on multi-part objects as segmentation noise propagates to part boundaries, most strongly in the Chamfer Distances; in aggregate \ArtMesh{} remains far more accurate on axis and motion metrics than GaussianArt given ground-truth masks. With predicted depth the impact is moderate; predicted poses dominate the remaining error.
\begin{table}[t]
\centering
\caption{Robustness to estimated depth and pose on the StorageFurniture objects of Articulate-100, scored as in Table~\ref{tab:articulate100}.}
\label{tab:robust-depthpose}
\setlength{\tabcolsep}{4pt}
\resizebox{\linewidth}{!}{\begin{tabular}{lccccc}
\toprule
 & Axis Ang $\downarrow$ & Axis Pos $\downarrow$ & Part Motion $\downarrow$ & CD-s $\downarrow$ & CD-m $\downarrow$ \\
\midrule
Ours                        & 3.29  & 0.03 & 5.16 & 22.14 & 17.64 \\
Ours (PredDepth)            & 6.07  & 0.48 & 7.59 & 25.73 & 23.81 \\
Ours (PredPose)             & 15.84 & 1.05 & 10.44 & 33.79 & 41.55 \\
Ours (PredDepth+PredPose)   & 15.63 & 1.10 & 13.29 & 45.21 & 59.33 \\
\bottomrule
\end{tabular}}
\end{table}

\textbf{Predicted part masks with PArtSAM.}
\label{app:partsam}
PArtSAM adapts the promptable 3D part segmentation model PartSAM~\cite{partsam} to our two-state setting. Each state's posed RGB-D views are back-projected into a colored point cloud with two per-point motion features, the nearest-neighbor displacement toward the other state and a binarized movability indicator, which enter through two added input channels whose first-layer weights are zero-initialized, so finetuning starts from the released model. We finetune it on a disjoint set of PartNet-Mobility objects from the benchmark categories, using their part annotations, to separate movable from static surfaces and to split the movable region into the given number of parts. Segmentation runs on the open state, the labels are carried to the closed state by per-part rigid ICP, and the point labels are rendered into every training view as segmentation maps.

\textbf{Reconstruction from predicted masks.}
Because predicted boundaries are less reliable than ground-truth ones, Ours-PArtSAM uses a modified schedule on all objects. The predicted maps only initialize the part labels: the mesh stays an unconnected triangle soup with trainable labels during articulation learning, each part's joint type is selected among revolute, prismatic and static, and the per-part remeshing and a final geometry and appearance refinement run after the articulation is learned. No ground-truth annotation of an evaluated object enters this route.

\textbf{Baseline under the same predicted masks.}
To compare methods under identical segmentation, we also train GaussianArt on the same PArtSAM-predicted masks for all 100 objects (Table~\ref{tab:robust-mask-partsam}). GaussianArt degrades from 39.15$^\circ$ to 47.04$^\circ$ Axis Angle and its movable-part Chamfer Distance rises by about half (57.45 to 85.19\,mm), whereas \ArtMesh{} rises only to 12.26$^\circ$ with Part Motion of 4.54 (3.64 with ground-truth masks), remaining about four times more accurate on Axis Angle and nine times on Part Motion than GaussianArt given the same masks.
\begin{table}[t]
\centering
\caption{Ground-truth vs.\ PArtSAM-predicted part masks on Articulate-100 (100 objects), with the same predicted masks given to both methods.}
\label{tab:robust-mask-partsam}
\setlength{\tabcolsep}{4pt}
\resizebox{\linewidth}{!}{\begin{tabular}{lcccc}
\toprule
 & Axis Ang $\downarrow$ & Part Motion $\downarrow$ & CD-s $\downarrow$ & CD-m $\downarrow$ \\
\midrule
GaussianArt (GT Mask)      & 39.15 & 29.32 & 25.96 & 57.45 \\
GaussianArt (PArtSAM)      & 47.04 & 41.97 & 28.77 & 85.19 \\
Ours (GT Mask)             & \textbf{3.49} & \textbf{3.64} & \textbf{18.98} & \textbf{15.01} \\
Ours (PArtSAM)             & 12.26 & 4.54 & 21.60 & 60.69 \\
\bottomrule
\end{tabular}}
\end{table}

\textbf{Novel-view rendering quality.}
Table~\ref{tab:novel-view} reports novel-view metrics on 20 held-out views per object in Articulate-100. \ArtMesh{} comes within 0.7\,dB of ArtGS's PSNR while its movable-part Chamfer Distance is about a quarter of ArtGS's.
\begin{table}[t]
\centering
\caption{Novel-view quality on 20 held-out views per object in Articulate-100. All methods use ground-truth part masks (ArtGS initializes its part centers from them).}
\label{tab:novel-view}
\setlength{\tabcolsep}{6pt}
\begin{tabular}{lcc}
\toprule
 & PSNR $\uparrow$ & CD-m $\downarrow$ \\
\midrule
ArtGS       & \textbf{42.35} &  58.66 \\
GaussianArt & 38.65 &  57.45 \\
Ours        & 41.66 &  \textbf{15.01} \\
\bottomrule
\end{tabular}
\end{table}

\textbf{Temporal consistency.}
Table~\ref{tab:tlpips} reports TLPIPS of \ArtMesh{} renderings across articulation states, matching the ground-truth level and demonstrating temporally consistent results. TLPIPS is the mean LPIPS (VGG) between consecutive frames of a video rendered from one fixed test camera while the joints sweep from the start to the end state and back, with 30 frames per direction. Ours drives the reconstruction with the predicted joints, and GT drives the same reconstruction with the ground-truth joints, so the comparison isolates temporal artifacts introduced by the predicted motion.
\begin{table}[t]
\centering
\caption{Temporal consistency (TLPIPS~$\downarrow$) of \ArtMesh{} results against ground truth.}
\label{tab:tlpips}
\setlength{\tabcolsep}{6pt}
\begin{tabular}{lcc}
\toprule
 & Ours & GT \\
\midrule
TLPIPS $\downarrow$ & 0.0076 & 0.0075 \\
\bottomrule
\end{tabular}
\end{table}

\textbf{Finer ablation and backbone comparison.}
Table~\ref{tab:finer-ablation} isolates the recovery stage after remeshing and the pixel-wise consistency loss. We also substitute Gaussian reconstruction backbones (3DGS as used by GaussianArt, and 2DGS~\cite{2dgs}) for our mesh backbone: both degrade joints, motion and geometry, raising Axis Ang above $49^\circ$, CD-s by about a third and CD-m to about twice that of the full model, showing the mesh-native backbone is better suited to articulation-aware optimization.
\begin{table}[t]
\centering
\caption{Finer ablation: recovery stage, pixel-wise consistency, and Gaussian reconstruction backbones substituted for the mesh backbone, scored as in Table~\ref{tab:articulate100}.}
\label{tab:finer-ablation}
\setlength{\tabcolsep}{4pt}
\resizebox{\linewidth}{!}{\begin{tabular}{lccccc}
\toprule
 & Axis Ang $\downarrow$ & Axis Pos $\downarrow$ & Part Motion $\downarrow$ & CD-s $\downarrow$ & CD-m $\downarrow$ \\
\midrule
Full                   & 3.49  & 0.03 & 3.64  & 18.98 & 15.01 \\
$-$Recovery            & 9.00  & 0.66 & 9.14  & 19.91 & 19.64 \\
$-$PixelLoss           & 46.64 & 2.15 & 27.38 & 19.04 & 31.12 \\
GaussianArt (Backbone) & 49.49 & 0.12 & 10.90 & 25.27 & 30.05 \\
2DGS (Backbone)        & 49.86 & 0.12 & 13.01 & 27.56 & 33.41 \\
\bottomrule
\end{tabular}}
\end{table}

\textbf{Interaction in SAPIEN.}
We load every reconstruction into SAPIEN with a fixed base, using a CoACD convex decomposition of each part as collision geometry and keeping all collision pairs enabled. With the joint drives off, we push each joint open with a constant force and count the joints that open over at least 90\% of their range (Table~\ref{tab:sim-sapien}). Each link has unit mass, the force is 20\,N for prismatic and 5\,N$\cdot$m for revolute joints, applied for 3\,s at 240\,Hz with joint damping 1, and CoACD uses a concavity threshold of 0.05 after quadric decimation. All of our exports load and simulate stably. Every method is driven by its own predicted joint motion with the joint range taken from the ground truth. Our directly exported mesh, with no TSDF step, opens more joints than the baselines' TSDF meshes, while all reconstructions remain below the ground-truth assets. The direct export is not guaranteed to be watertight, so we recommend the TSDF pass when watertight collision geometry is required.
\begin{table}[t]
\centering
\caption{Interaction success in SAPIEN on Articulate-100: fraction of joints that a constant force (20\,N prismatic, 5\,N\,m revolute) opens over at least 90\% of their range. Collision geometry is a CoACD convex decomposition of each part with all collision pairs enabled. Each method is driven by its own predicted joint motion (axis, pivot and direction from the part paired to each ground-truth joint, joint type as in Table~\ref{tab:articulate100}), with the joint range taken from the ground truth. Baselines use their TSDF meshes; ours uses the directly exported mesh. A joint whose predicted type is wrong or that has no paired part, and every joint of a reconstruction that fails to load, counts as not opened. GT: PartNet-Mobility meshes with ground-truth joints.}
\label{tab:sim-sapien}
\setlength{\tabcolsep}{4pt}
\resizebox{\linewidth}{!}{\begin{tabular}{lccccc}
\toprule
 & GT & GaussianArt + TSDF & ArtGS + TSDF & ReArtGS++ + TSDF & Ours \\
\midrule
Pull-open success (\%) $\uparrow$ & 91.3 & 39.3 & 67.2 & 65.6 & \textbf{74.4} \\
\bottomrule
\end{tabular}}
\end{table}

\section{Baseline Failure Analysis on Articulate-100}
\label{app:baseline-failure}
The failure patterns of ArtGS and GaussianArt match the limitations each paper identifies, amplified by our broader category mix. ArtGS uses center-based part assignment initialized by spectral clustering over Gaussians flagged ``dynamic'' via a Chamfer-distance threshold~\cite{artgs}; this degrades when parts share motion direction or overlap spatially, the case for multi-drawer storage and multi-leaf tables, which dominate our 3+part buckets. GaussianArt, even with correct part labels, lacks part-aware motion learning: its per-Gaussian transform is a weighted blend of per-part motion bases, hardened by $\arg\max$ over the blending weights after a soft stage~\cite{gaussianart}, so when movable parts are adjacent and visually similar to the static body, the blended motion can lock in an axis belonging to neither. ReArtGS++ is the strongest baseline on two-part objects, but its Axis Ang and movable-part Chamfer Distance grow with the number of parts (Table~\ref{tab:articulate100}): its distance-based soft part assignment has only a weak local-voting term at part boundaries, so adjacent movable parts with similar appearance can still share Gaussians and each part's motion is fit to a mixed set. \ArtMesh{} avoids this through bidirectional image, color, and opacity consistency losses applied to locked per-state meshes after per-part restricted Delaunay (Sec. ~\ref{sec:part_aware_mesh_fields}).

\section{Failure Case Analysis and Limitations}
\label{app:failure}

\begin{figure}[t]
    \centering
    \includegraphics[width=\linewidth]{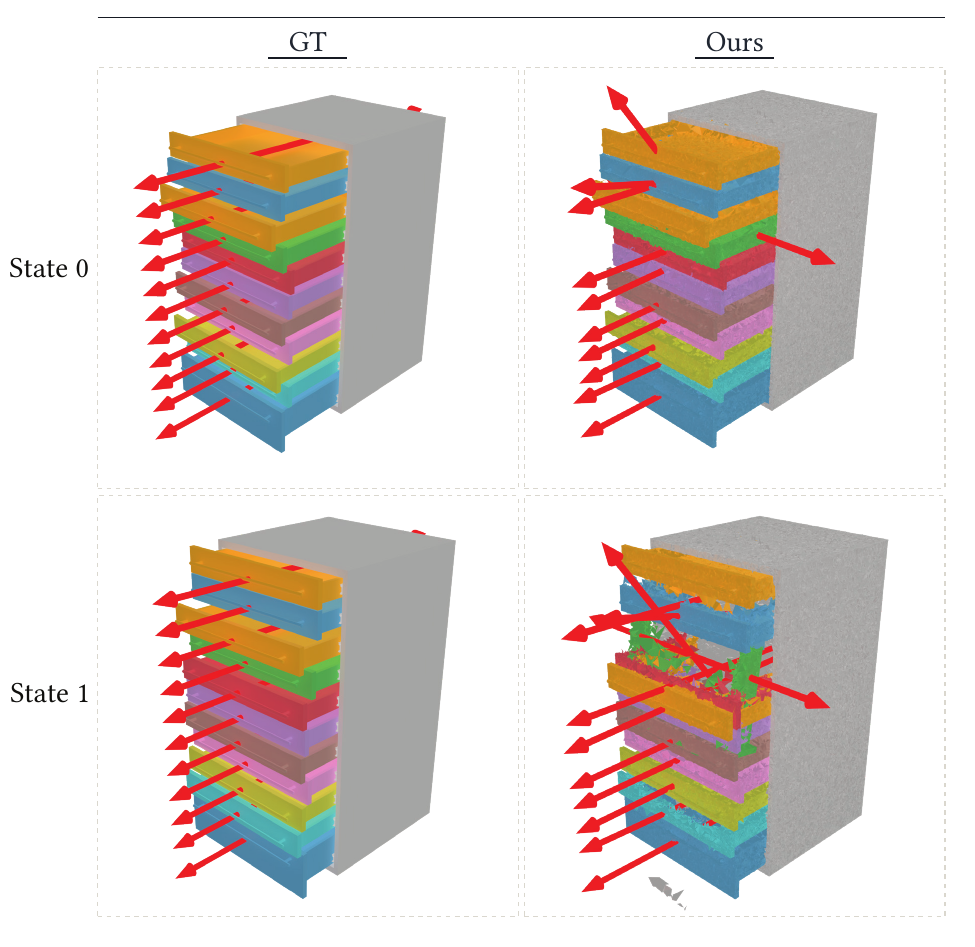}
    \caption{Failure case under heavy occlusion. When a movable part is largely hidden in both states, \ArtMesh{} can recover an inaccurate motion axis or fail to separate the part cleanly.}
    \label{fig:failure}
\end{figure}
When a part is visible from only a small number of views in either state, this evidence becomes sparse and our optimization is correspondingly under-constrained. Two failure modes arise in practice. First, a movable part that is largely occluded in both states (for instance, a small drawer between two other drawers with similar motion patterns) leaves the consistency losses with too few corresponded vertices to reliably estimate its rigid motion; the recovered axis can drift in direction or position by a noticeable margin. Second, when the occluded part's neighbors share similar motion patterns and appearance, the part may fail to separate cleanly from them, blending into an adjacent movable component. Figure~\ref{fig:failure} shows a representative case. Both modes come from sparse observations. Denser viewpoints around occluded regions, or articulation states that vary for each movable part, would resolve them; we leave input acquisition for articulated reconstruction to future work.

\section{Training Details}
\label{app:training}

\textbf{Scheduling details.} We train each object for a total of $T=4\!\times\!10^{4}$ iterations,
evenly split into a reconstruction phase ($s_1=2\!\times\!10^{4}$) and an articulation phase ($s_2=2\!\times\!10^{4}$). The mesh is initialized by back-projecting the RGB-D observations into a per-state point cloud and seeding one small triangle per point (a triangle soup), which reconstruction then optimizes and the restricted-Delaunay step later connects into per-part meshes. The reconstruction phase optimizes the meshes $\mathcal{M}(t_1)$ and $\mathcal{M}(t_2)$ at both states $t_1,t_2\in\{0,1\}$ and their attributes under the per-state image loss only; densification and the two upsampling steps both complete with at least $2000$ iterations of adaptation time before the phase boundary to leave the mesh in a stable state. At iteration $1.5\!\times\!10^{4}$ of the reconstruction phase we harden the part affinities via argmax and run the per-part restricted Delaunay step \cite{meshsplatting}, followed by a short reconstruction stage with fixed topology that refines vertex positions and appearance after remeshing. At the phase boundary we then freeze the canonical vertex positions and topology and initialize the articulation field: per-part axis directions of $R_k^{+}$ are seeded from PCA of each movable part's vertices, and pivots $P_k^{+}$ are computed from part geometry. Because a revolute pivot can slide along the rotation axis without changing the motion, Axis Pos measures the distance between the predicted and ground-truth axis lines rather than between pivots. Appearance uses degree-3 spherical harmonics; the cross-state vertex-wise consistency compares only the view-independent (DC) coefficient of $c_i$, and part rotations are applied to vertex positions without rotating the higher-order SH coefficients, an approximation accurate for the largely diffuse benchmark objects.
The articulation phase schedules the four cycle components relative to the articulation step as follows. The forward pixel-wise motion-consistency loss $\mathcal{L}_{\mathrm{pix}}^{1\rightarrow2}$ is active from the first articulation step ($2\!\times\!10^{4}$). The
backward pixel-wise motion-consistency loss $\mathcal{L}_{\mathrm{pix}}^{2\rightarrow1}$ is delayed to iteration $2.5\!\times\!10^{4}$, which
lets the forward direction commit to an axis before the inverse transport starts enforcing agreement. At the beginning of the second half of articulation stage we resolve the rotation-vs-translation type decision (discussed in joint-type handling below), and activate the vertex-wise half of the consistency objective: $\mathcal{L}_{\mathrm{vtx}}^{1\rightarrow2}$ and $\mathcal{L}_{\mathrm{vtx}}^{2\rightarrow1}$.

\textbf{Loss weights.} The motion-consistency weights are
$\lambda_{\mathrm{VMC}}=\lambda_{\mathrm{PMC}}=5\!\times\!10^{-2}$ in all our experiments. The
per-pixel image losses use $\lambda_{\mathrm{rgb}}=1-\lambda_{\mathrm{ssim}}$ with
$\lambda_{\mathrm{ssim}}=0.2$, matching the reconstruction-phase image loss. The reconstruction-phase loss in Eq.~\eqref{eq:state_reconstruction_loss} uses $\lambda_{\mathrm{rgb}}=0.8$, $\lambda_{\mathrm{ssim}}=0.2$, $\lambda_{\mathrm{depth}}=0.5$, $\lambda_{\mathrm{part}}=0.5$. Within $\mathcal{L}_{\mathrm{vtx}}$, $\lambda_x=\lambda_c=\lambda_\sigma=1$; each iteration evaluates 256 randomly sampled source vertices per direction, and $\eta_i$ restricts correspondences to target vertices of the same part. The reconstruction phase also uses an opacity-entropy term (weight 0.01), a background-alpha term (0.5), normal regularization ($5\!\times\!10^{-5}$ and $10^{-2}$, from iteration 5000) and a vertex-depth term ($2.5\!\times\!10^{-4}$, from iteration 12000).

\textbf{Learning rates.} Articulation uses constant (non-decayed) per-group
learning rates throughout the articulation phase: $\eta_{R}=8\!\times\!10^{-3}$ for rotation parameters of $R_k^{+}$ (axis direction
and magnitude), $\eta_{T}=1\!\times\!10^{-3}$ for translation $T_k^{+}$, and $0.1\,\eta_{T}=1\!\times\!10^{-4}$ for joint pivots $P_k^{+}$. The part-weight logits $s_i$ use $\eta_{w}=5\!\times\!10^{-3}$ while soft. Per-vertex SH coefficients $c_i(t)$ use $\eta_{c}=1.6\!\times\!10^{-3}$ and per-vertex opacity $\sigma_i(t)$ uses $\eta_{\sigma}=3\!\times\!10^{-2}$, both inherited from the reconstruction
phase; vertex positions $x_i(t)$ decay exponentially from
$2\!\times\!10^{-4}$ to $2\!\times\!10^{-6}$ over the reconstruction phase and are frozen thereafter.

\textbf{Joint-type handling.} We do not assume the joint type at the start of training. The decision between revolute and prismatic is resolved via a head-to-head competition during the first half of the articulation phase. At the start of articulation we clone the per-part articulation parameters into two parallel candidate sets: a \emph{revolute candidate} $(R_k^{+,\text{rot}},P_k^{+,\text{rot}})$ with $T_k^{+,\text{rot}}=0$, and a \emph{prismatic candidate} $T_k^{+,\text{trans}}$ with $R_k^{+,\text{trans}}=I$ and no pivot. Each candidate set has its own optimizer and is updated on alternating iterations: even iterations render the mesh $\mathcal{M}(t_1)$ articulated to $\mathcal{M}(t_2)$ using only the revolute candidate (Eq.~\eqref{eq:forward} with $T_k^{+}=0$), odd iterations render using only the prismatic candidate ($R_k^{+}=I$), and in both cases supervision comes from the standard photometric loss $\mathcal{L}_{\mathrm{pix}}^{1\rightarrow2}$ against state-$t_2$ ground truth. The main articulation parameters $(R_k^{+},T_k^{+},P_k^{+})$ remain frozen at their initialization during this alternation phase, waiting to be replaced by the winning candidate. Quaternions are renormalized after each rotation step. At the midpoint of the articulation phase we run a per-part bake-off on state-$t_2$ training cameras: for each part $k$, we render once with only the revolute candidate active on part $k$ (all other parts identity) and once with only the prismatic candidate active on part $k$, mask the photometric loss by the state-$t_2$ semantic map restricted to part $k$, and average across end-state views to obtain $\mathcal{L}^{\text{rot}}_k$ and $\mathcal{L}^{\text{trans}}_k$. The masked, per-part comparison gives a cleaner signal than a global loss would, since each part is judged only on the pixels it actually owns. Part $k$ is classified as revolute if $\mathcal{L}^{\text{rot}}_k \le \mathcal{L}^{\text{trans}}_k$ and prismatic otherwise; the winning candidate is merged into the main articulation field and the Adam moments of $(R_k^{+},T_k^{+},P_k^{+})$ are reset to zero so that the subsequent unified optimization starts from a clean second-moment estimate. From this point on, the constraints described above (zeroed $T_k^{+}$ for revolute parts, $R_k^{+}=I$ and $P_k^{+}=0$ for prismatic parts) are enforced by masking the corresponding gradients before each optimizer step. After the parallel joint type optimization, for joints classified as prismatic, their rotation $R_k^{+}$ is set to identity and their pivot $P_k^{+}$ is zeroed and frozen, and optimization continues
only on $T_k^{+}$. For parts classified
as revolute, their translation $T_k^{+}$ is zeroed and frozen, and optimization continues only on $(R_k^{+},P_k^{+})$. This thresholding constrains each
joint to behave as exactly one kinematic primitive and prevents the articulation field from oscillating between rotation and translation explanations during the remainder of training.

\textbf{State-wise mesh reconstruction.}
 Before estimating articulation, each state mesh is optimized against its own observations. The reconstruction objective is
\begin{equation}
\mathcal{L}_{\mathrm{rec}}
=
\sum_{t\in\{t_1,t_2\}}
\frac{1}{|\mathcal{C}_t|}
\sum_{m\in\mathcal{C}_t}
\ell_{\mathrm{rec}}
\left(
\hat{O}_m(t);
O_m(t)
\right),
\label{eq:state_reconstruction_objective}
\end{equation}
where $\mathcal{C}_t$ is the set of training cameras at state $t$, $m$ indexes a camera in $\mathcal{C}_t$, $O_m(t)=(I_m(t),D_m(t),S_m(t))$ collects the observed image, depth and part map, and $\hat{O}_m(t)$ the rendered ones. The per-view loss is
\begin{equation}
\begin{aligned}
\ell_{\mathrm{rec}}
=
&
\lambda_{\mathrm{rgb}}
\left\|
\hat{I}-I
\right\|_1
+
\lambda_{\mathrm{ssim}}
\left(
1-\mathrm{SSIM}(\hat{I},I)
\right)
\\
&
+
\lambda_{\mathrm{depth}}
\left\|
\hat{D}-D
\right\|_{1,\Omega_D}
+
\lambda_{\mathrm{part}}
\mathrm{CE}(\hat{S},S).
\end{aligned}
\label{eq:state_reconstruction_loss}
\end{equation}
Here $\Omega_D$ denotes valid depth pixels, and $S$ is the semantic part map. The part-mask term supervises the part logits $s_i(t)$  whose hardened labels $p_i(t)$ later define the per-part mesh topology.

\textbf{Hardware and runtime.} Each object is trained on a single NVIDIA A10 GPU. Each run takes approximately 30 minutes of wall-clock time end-to-end, varying with total part number in each object.

\textbf{Part masks.} In the main experiments \ArtMesh{} and GaussianArt, the two methods that consume part masks, use ground-truth labels, and ArtGS and ReArtGS++, which take no part masks, have their part centers initialized from them.

\section{Additional Method Details}
\label{app:method-details}

\textbf{Homogeneous form and analytic inverse.}
We give the full derivation of the analytic inverse motion summarized in Sec.~\ref{method:consistency}. The forward motion $g_k^{+}$ of Eq.~\eqref{eq:forward} has translation
\begin{equation}
b_k^{+} \;=\; -R_{k}^{+}P_{k}^{+} + P_{k}^{+} + T_{k}^{+}.
\label{eq:forward_translation}
\end{equation}
Equivalently, in homogeneous coordinates,
\begin{equation}
G_k^+
=
\begin{bmatrix}
R_k^+ & b_k^+ \\
0 & 1
\end{bmatrix},
\qquad
\bar{x}
=
\begin{bmatrix}
x\\
1
\end{bmatrix}.
\label{eq:homogeneous_forward_motion}
\end{equation}
The backward motion $(R_k^{-},T_k^{-},P_k^{-})$ that maps state $t_2$ back to state $t_1$ is not independently learned. It is the analytic inverse of the forward affine transform:
\begin{equation}
G_k^- = (G_k^+)^{-1}
=
\begin{bmatrix}
(R_k^+)^\top & -(R_k^+)^\top b_k^+ \\
0 & 1
\end{bmatrix}.
\label{eq:homogeneous_inverse_motion}
\end{equation}
Therefore,
\begin{equation}
g_k^-(y)
=
(R_k^+)^\top(y-b_k^+)
=
R_k^-y+b_k^- ,
\label{eq:inverse_affine_motion}
\end{equation}
with
\begin{equation}
R_k^-=(R_k^+)^\top,
\qquad
b_k^-=-(R_k^+)^\top b_k^+ .
\label{eq:inverse_affine_parameters}
\end{equation}
If written in pivot form with $P_k^- = P_k^+$, the inverse translation is
\begin{equation}
T_k^- = -(R_k^+)^\top T_k^+ .
\label{eq:inverse_pivot_translation}
\end{equation}

\textbf{Differentiable rendering.}
We detail the alpha compositing used by the renderer $\mathcal{R}$ in Eq.~\eqref{eq:mesh_renderer_outputs}. For RGB rendering, faces overlapping a pixel are ordered front-to-back (Fig.~\ref{fig:render}(b)). Let $f_1,\ldots,f_N$ be the ordered faces contributing to pixel $\mathbf{p}$, and let
\begin{equation}
\alpha_n(\mathbf{p})
=
\sigma_n \phi_n(\mathbf{p})
, \qquad \phi_n(\mathbf{p}) =
\left(\mathrm{ReLU}\!\left(
\frac{\psi_n(\mathbf{p})}{\psi_n(\mathbf{s}_n)}
\right)
\right)^{\gamma}
\label{eq:face_alpha}
\end{equation}
denote the contribution of face $f_n$, where $\phi_n(\mathbf{p})$ is the smooth triangle window function, $\psi_n$ is the signed distance field of the projected triangle $f_n$, $\mathbf{s}_n$ is its incenter, $\gamma$ is a sharpness exponent, and $\sigma_n$ is the face opacity obtained from its vertices. The rendered color is accumulated by alpha compositing:
\begin{equation}
\hat{I}(\mathbf{p})
=
\sum_{n=1}^{N}
c_n
\alpha_n(\mathbf{p})
\prod_{r=1}^{n-1}
\left(
1-\alpha_r(\mathbf{p})
\right),
\label{eq:alpha_compositing}
\end{equation}
where $c_n$ is the view-dependent face color obtained by barycentric interpolation of the vertex color coefficients. Depth and part logits are accumulated using the same transmittance weights.

\section{Additional Baselines}
\label{app:extra-baselines}

We further compare against DigitalTwinArt~\cite{digitaltwinart} (DTA), Part2GS~\cite{part2gs}, and SplArt~\cite{splart} on Articulate-100 (all 100 objects; SplArt, which models a single moving part, on the 34 two-part objects), each without and with GT part masks (Table~\ref{tab:extra-baselines}), using the matching, joint-type handling and failure penalties of Table~\ref{tab:articulate100}. Without GT masks, each baseline runs its released code and configuration, with the Part2GS changes described below. With GT masks, we follow the GT-mask variants of ArtGS and ReArtGS++ in Table~\ref{tab:articulate100}: the masks only initialize each method's own part assignment, and all later optimization is unchanged. Each point of the method's own reconstruction is projected into the GT part masks of 60 training views and takes the majority label over the views where it is visible (depth tolerance 2\,cm); points never visible keep the method's own initialization. These labels initialize DTA's per-state part classifier, Part2GS's part centers and part identities, and SplArt's per-Gaussian mobility. All methods are given the true number of parts.

\begin{table*}[t]
\centering
\caption{Additional baselines on Articulate-100. Left: methods given GT part masks, which initialize each method's own part assignment (Sec.~\ref{app:extra-baselines}); all later optimization is unchanged. Right: methods without GT masks (Ours-PArtSAM uses PArtSAM masks; the baselines run as released, Part2GS with the changes described in Sec.~\ref{app:extra-baselines}). Matching, joint-type handling and failure penalties are identical to Table~\ref{tab:articulate100}. \textbf{Best} per column within each half in bold. SplArt models a single moving part, so it is reported on two-part objects only (-). Gain is the improvement of Ours over the best baseline in the same half.}
\label{tab:extra-baselines}
\setlength{\tabcolsep}{3pt}
\renewcommand{\arraystretch}{0.85}
\resizebox{\linewidth}{!}{
\begin{tabular}{ll rrrrr @{\hskip 10pt} l rrrrr}
\toprule
 & \multicolumn{6}{c}{\textbf{With GT masks}} & \multicolumn{6}{c}{\textbf{Without GT masks}} \\
\cmidrule(lr){2-7}\cmidrule(lr){8-13}
 & Method & 2 (34) & 3 (20) & 4-5 (31) & 6+ (15) & All (100) & Method & 2 (34) & 3 (20) & 4-5 (31) & 6+ (15) & All (100) \\
\midrule
Axis Ang~($^\circ$) $\downarrow$ & DTA-GTMask & 3.11 & 7.48 & 11.90 & 18.59 & 9.03 & DTA & \textbf{0.88} & 14.84 & 24.58 & 32.78 & 15.80 \\
 & Part2GS-GTMask & 8.50 & 9.18 & 11.88 & 22.29 & 11.75 & Part2GS & 6.16 & 16.53 & 22.77 & 28.84 & 16.78 \\
 & SplArt-GTMask & \textbf{0.12} & - & - & - & - & SplArt & 2.81 & - & - & - & - \\
 & \our{Ours-GTMask} & \our{2.02} & \our{\textbf{4.30}} & \our{\textbf{2.93}} & \our{\textbf{6.90}} & \our{\textbf{3.49}} & \our{Ours-PArtSAM} & \our{1.68} & \our{\textbf{12.02}} & \our{\textbf{19.79}} & \our{\textbf{20.97}} & \our{\textbf{12.26}} \\
\rowcolor{gray!12}  & Gain & $-$1.90 & +3.18 & +8.95 & +11.69 & +5.54 & Gain & $-$0.80 & +2.82 & +2.98 & +7.87 & +3.54 \\
\midrule
Axis Pos~(0.1\,m) $\downarrow$ & DTA-GTMask & 2.88 & 1.80 & 0.55 & 7.22 & 2.59 & DTA & 0.13 & 1.44 & 1.65 & 3.49 & 1.37 \\
 & Part2GS-GTMask & 1.07 & 1.00 & 1.17 & 1.97 & 1.22 & Part2GS & 0.75 & 2.29 & 1.57 & 2.31 & 1.54 \\
 & SplArt-GTMask & \textbf{0.00} & - & - & - & - & SplArt & \textbf{0.01} & - & - & - & - \\
 & \our{Ours-GTMask} & \our{0.04} & \our{\textbf{0.03}} & \our{\textbf{0.01}} & \our{\textbf{0.03}} & \our{\textbf{0.03}} & \our{Ours-PArtSAM} & \our{0.06} & \our{\textbf{0.34}} & \our{\textbf{0.46}} & \our{\textbf{0.72}} & \our{\textbf{0.34}} \\
\rowcolor{gray!12}  & Gain & $-$0.04 & +0.97 & +0.54 & +1.94 & +1.19 & Gain & $-$0.05 & +1.10 & +1.11 & +1.59 & +1.03 \\
\midrule
Part Motion~($^\circ$/m) $\downarrow$ & DTA-GTMask & 11.87 & 5.31 & 2.39 & 6.04 & 6.75 & DTA & 5.07 & 12.16 & 8.78 & 18.02 & 9.58 \\
 & Part2GS-GTMask & 10.03 & 9.05 & 6.47 & 12.64 & 9.12 & Part2GS & 7.51 & 13.92 & 11.73 & 19.78 & 11.94 \\
 & SplArt-GTMask & \textbf{5.37} & - & - & - & - & SplArt & 5.41 & - & - & - & - \\
 & \our{Ours-GTMask} & \our{5.82} & \our{\textbf{2.12}} & \our{\textbf{2.32}} & \our{\textbf{3.43}} & \our{\textbf{3.64}} & \our{Ours-PArtSAM} & \our{\textbf{1.26}} & \our{\textbf{3.60}} & \our{\textbf{6.68}} & \our{\textbf{8.79}} & \our{\textbf{4.54}} \\
\rowcolor{gray!12}  & Gain & $-$0.45 & +3.19 & +0.07 & +2.61 & +3.11 & Gain & +3.81 & +8.56 & +2.10 & +9.23 & +5.04 \\
\midrule
CD-s~(mm) $\downarrow$ & DTA-GTMask & \textbf{17.67} & 19.07 & 20.19 & 22.06 & 19.39 & DTA & \textbf{16.57} & 21.45 & \textbf{21.43} & 22.20 & \textbf{19.90} \\
 & Part2GS-GTMask & 21.63 & 20.37 & 26.32 & 26.47 & 23.56 & Part2GS & 19.23 & 21.89 & 24.79 & 28.58 & 22.89 \\
 & SplArt-GTMask & 25.12 & - & - & - & - & SplArt & 25.45 & - & - & - & - \\
 & \our{Ours-GTMask} & \our{17.95} & \our{\textbf{18.86}} & \our{\textbf{19.66}} & \our{\textbf{20.11}} & \our{\textbf{18.98}} & \our{Ours-PArtSAM} & \our{18.60} & \our{\textbf{19.62}} & \our{26.06} & \our{\textbf{21.79}} & \our{21.60} \\
\rowcolor{gray!12}  & Gain & $-$0.28 & +0.21 & +0.53 & +1.95 & +0.41 & Gain & $-$2.03 & +1.83 & $-$4.63 & +0.41 & $-$1.70 \\
\midrule
CD-m~(mm) $\downarrow$ & DTA-GTMask & 29.62 & 36.54 & 64.41 & 75.02 & 48.60 & DTA & 18.43 & 155.91 & 176.51 & 222.25 & 125.50 \\
 & Part2GS-GTMask & 35.28 & 49.62 & 109.74 & 161.18 & 80.12 & Part2GS & 36.93 & 193.06 & 204.77 & 172.95 & 140.59 \\
 & SplArt-GTMask & 19.89 & - & - & - & - & SplArt & 26.60 & - & - & - & - \\
 & \our{Ours-GTMask} & \our{\textbf{11.09}} & \our{\textbf{15.74}} & \our{\textbf{15.27}} & \our{\textbf{22.40}} & \our{\textbf{15.01}} & \our{Ours-PArtSAM} & \our{\textbf{15.13}} & \our{\textbf{79.87}} & \our{\textbf{85.67}} & \our{\textbf{86.78}} & \our{\textbf{60.69}} \\
\rowcolor{gray!12}  & Gain & +8.80 & +20.80 & +49.14 & +52.62 & +33.59 & Gain & +3.30 & +76.04 & +90.84 & +86.17 & +64.81 \\
\bottomrule
\end{tabular}
}
\end{table*}

\begin{figure*}[t]
    \centering
    \includegraphics[width=0.8\linewidth]{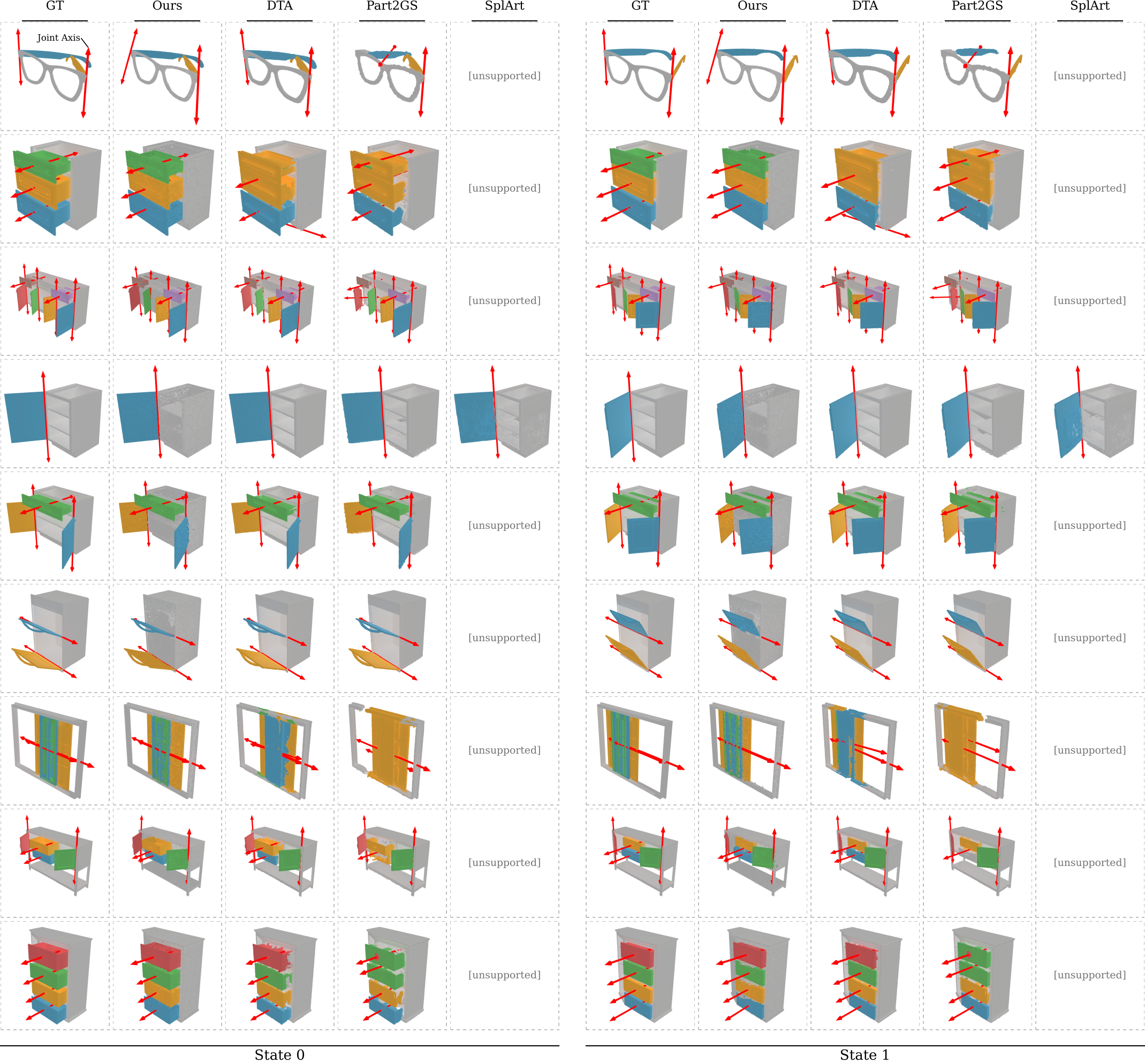}
    \caption{Qualitative comparisons with the additional baselines on the objects of Fig.~\ref{fig:articulate100-qual-full} at both states. All methods use ground-truth part masks. SplArt models a single moving part and is shown only for the object with one moving part. DTA and Part2GS merge adjacent drawers and window sashes into one part.}
    \label{fig:extra-baselines-qual}
\end{figure*}

With GT masks, Ours-GTMask has the lowest error on every metric over all 100 objects, with a CD-m of 15.01\,mm against 48.60\,mm for DTA-GTMask and 80.12\,mm for Part2GS-GTMask. Without GT masks, Ours-PArtSAM has the lowest joint errors and CD-m (60.69\,mm against 125.50\,mm for DTA and 140.59\,mm for Part2GS), while DTA, which reconstructs a signed distance field from RGB-D, reaches a lower CD-s. The margin grows with the number of parts: on objects with six or more parts, the CD-m of Ours-PArtSAM is 86.78\,mm against 222.25\,mm for DTA and 172.95\,mm for Part2GS. SplArt models a single moving part, so we report it on two-part objects only, where it is competitive; DTA reports experiments with up to three parts. Fig.~\ref{fig:extra-baselines-qual} compares the reconstructions on the objects of Fig.~\ref{fig:articulate100-qual-full}.

\textbf{Part2GS implementation.}
The released Part2GS code evaluates four losses over all pairs of Gaussians at every iteration (repulsion, neighborhood part consistency, contact, and vector-field alignment), which exceeds 80\,GB of GPU memory once densification reaches a few $10^4$ Gaussians. Following the paper's definition of the part loss over the Gaussians in the current batch, we evaluate these four losses on a random batch of 8192 Gaussians per iteration, with nearest neighbors still searched over all Gaussians. The released code also assigns the initial part identities from the ground-truth joint parameters and selects its output checkpoint by the joint error against ground truth; we instead use the clustering-based part assignment included in the same code and the final checkpoint, so that no ground-truth joint information enters training or model selection. We further fix two crashes (the per-Gaussian part data is not extended during densification, and a visualization indexes at most ten parts) and evaluate each state's deformation once per iteration. One run takes 22 to 40 minutes on one A800 GPU.

\section{Additional Analyses}
\label{app:extra-analyses}

\textbf{Joint errors by joint type.}
Part Motion averages rotation errors in degrees for revolute joints and translation errors in meters for prismatic joints. Table~\ref{tab:joint-type-split} reports the two joint types separately. With GT masks, \ArtMesh{} has the lowest error on every metric for both types, with a revolute motion error of $8.48^\circ$ against $14.68^\circ$ for ReArtGS++-GTMask and a prismatic motion error of 0.028\,m against 0.050\,m for ArtGS-GTMask.

\begin{table}[t]
\centering
\caption{Joint errors on Articulate-100 split by joint type, with GT part masks. Errors are averaged over all 112 revolute and 175 prismatic ground-truth joints (per joint, whereas Table~\ref{tab:articulate100} averages per object), with the matching and failure penalties of Table~\ref{tab:articulate100}. Axis Pos applies to revolute joints only. \textbf{Best} per column in bold.}
\label{tab:joint-type-split}
\setlength{\tabcolsep}{3pt}
\resizebox{\linewidth}{!}{\begin{tabular}{lccccc}
\toprule
 & \multicolumn{3}{c}{Revolute (112)} & \multicolumn{2}{c}{Prismatic (175)} \\
\cmidrule(lr){2-4}\cmidrule(lr){5-6}
Method & Axis Ang ($^\circ$) & Axis Pos (0.1\,m) & Motion ($^\circ$) & Axis Ang ($^\circ$) & Motion (m) \\
\midrule
GaussianArt-GTMask & 47.01 & 4.75 & 49.66 & 27.07 & 0.320 \\
ArtGS-GTMask & 17.01 & 1.90 & 18.35 & 5.59 & 0.050 \\
ReArtGS++-GTMask & 7.53 & 1.31 & 14.68 & 16.41 & 0.190 \\
\our{Ours-GTMask} & \our{\textbf{4.72}} & \our{\textbf{0.06}} & \our{\textbf{8.48}} & \our{\textbf{4.97}} & \our{\textbf{0.028}} \\
\bottomrule
\end{tabular}}
\end{table}

\textbf{Effect of mask quality.}
The PArtSAM masks of Ours-PArtSAM have a mean movable-part mIoU of 0.72 over Articulate-100, decreasing from 0.93 on two-part objects to 0.56 on objects with six or more parts. Table~\ref{tab:mask-quality-subset} restricts the comparison without GT masks to the objects whose masks reach an mIoU of at least 0.5 or 0.7. On both subsets Ours-PArtSAM has the lowest error on all five metrics. Its errors track mask quality (Spearman $\rho=-0.64$ for Axis Ang and $-0.85$ for CD-m, $p<0.001$), and on the 53 objects with mIoU of at least 0.8 it reaches an Axis Ang of $2.51^\circ$, against $2.55^\circ$ for Ours-GTMask on the same objects. The gap between Ours-PArtSAM and Ours-GTMask in Table~\ref{tab:articulate100} is therefore strongly associated with segmentation errors on objects with many parts.

\begin{table}[t]
\centering
\caption{Methods without GT masks on the Articulate-100 objects whose PArtSAM movable-part masks reach a given quality (mean movable-part mIoU over both states), scored as in Table~\ref{tab:articulate100}. \textbf{Best} per column in bold.}
\label{tab:mask-quality-subset}
\setlength{\tabcolsep}{3pt}
\resizebox{\linewidth}{!}{\begin{tabular}{llccccc}
\toprule
Subset & Method & Axis Ang & Axis Pos & Part Motion & CD-s & CD-m \\
\midrule
\multirow{4}{*}{\shortstack{mIoU $\geq$ 0.5\\(75 objects)}} & GaussianArt-PArtSAM & 53.03 & 5.78 & 52.02 & 26.66 & 67.30 \\
 & ArtGS & 10.31 & 0.95 & 8.80 & 21.30 & 76.22 \\
 & ReArtGS++ & 9.17 & 0.68 & 8.24 & 30.18 & 83.47 \\
 & \our{Ours-PArtSAM} & \our{\textbf{7.77}} & \our{\textbf{0.23}} & \our{\textbf{3.79}} & \our{\textbf{19.33}} & \our{\textbf{33.11}} \\
\midrule
\multirow{4}{*}{\shortstack{mIoU $\geq$ 0.7\\(60 objects)}} & GaussianArt-PArtSAM & 55.97 & 6.31 & 56.75 & 25.84 & 62.57 \\
 & ArtGS & 8.62 & 0.86 & 7.56 & 19.98 & 60.72 \\
 & ReArtGS++ & 7.37 & 0.73 & 8.54 & 30.03 & 76.80 \\
 & \our{Ours-PArtSAM} & \our{\textbf{3.25}} & \our{\textbf{0.09}} & \our{\textbf{1.79}} & \our{\textbf{18.71}} & \our{\textbf{17.48}} \\
\bottomrule
\end{tabular}}
\end{table}

\end{document}